\documentclass[11pt]{article}    

\usepackage{verbatim}

\usepackage{fancybox}
\usepackage{amsmath}
\usepackage{amssymb}
\usepackage[dvips]{graphicx}
\usepackage{psfrag}
\usepackage{epic}
\usepackage{eepic}
\usepackage{natbib}
\usepackage{hyperref}

\usepackage{algpseudocode,algorithm,algorithmicx}
\newcommand*\Let[2]{\State #1 $\gets$ #2}

\DeclareGraphicsExtensions{ps}
\graphicspath{{fig/}{../fig/}{tmp/}}
\makeatletter
  \newcommand\figcaption{\def\@captype{figure}\caption}
  \newcommand\tabcaption{\def\@captype{table}\caption}
\makeatother

\def\mi{\begin{equation}}
\def\mf{\end{equation}}
\def\mia{\begin{eqnarray}}
\def\mfa{\end{eqnarray}}

\def\mean{\overline}
\def\exp{\mathcal{E}}
\def\reff#1{(\ref{#1})}

\newcommand{\ud}{\mathrm{d}}

\renewcommand{\v}[1]{\ensuremath{\mathbf{#1}}} 
\newcommand{\gv}[1]{\ensuremath{\mbox{\boldmath$ #1 $}}}  
\let\baraccent=\= 
\renewcommand{\=}[1]{\stackrel{#1}{=}} 

\def\mean{\mathcal{E}}

\newcommand{\transp}{\top}
\newcommand{\dynm}{\mathcal{M}}
\newcommand{\obsm}{\mathcal{H}}
\newcommand{\gaus}{\mathcal{N}}

\newcommand{\rkhs}{\mathcal{F}}

\newif\ifprintcallout
\printcallouttrue

\def\come#1{}
\usepackage{graphics,color}

\graphicspath{{./}}
\providecommand{\norm}[1]{\big\lVert#1\big\rVert}

\newcommand{\linelabel}[1]{}
\newcommand{\pj}[1]{{#1}} 
\newcommand{\ma}[1]{#1}

\title{Kernel embedding of maps for sequential Bayesian inference: \\The variational mapping particle filter}
\author{Manuel Pulido$^{a,b}$ and Peter Jan vanLeeuwen$^{a,c}$\\
  \scriptsize
$^a$ Data Assimilation Research Centre, Department of Meteorology, University of Reading, UK\\
\scriptsize$^b$ Department of Physics, FaCENA, Universidad Nacional del Nordeste and CONICET, Argentina\\
\scriptsize$^c$ National Centre for Earth Observation, Department of Meteorology, University of Reading, UK}

\begin{document}

\maketitle

\begin{abstract}
In this work, a novel sequential Monte Carlo filter is introduced which aims at efficient sampling of high-dimensional state spaces with a limited number of particles. 
Particles are pushed forward  from the \pj{prior to the posterior density using a sequence of mappings that minimizes the 
  Kullback-Leibler divergence between the posterior and the sequence of intermediate densities}.\ma{The sequence of mappings represents a gradient flow.}
A key ingredient of the mappings is that they are embedded in a reproducing kernel Hilbert space\pj{, which allows for a practical and efficient algorithm. The embedding provides a direct means to calculate the gradient of the Kullback-Leibler divergence leading to} quick convergence using well-known gradient-based stochastic optimization algorithms. Evaluation of the method is conducted in the \pj{chaotic} Lorenz-63 system,  \pj{the Lorenz-96 system, which is} a coarse prototype of atmospheric dynamics, and an epidemic model that describes cholera dynamics. No resampling is required in the mapping particle filter even for long recursive \ma{sequences}. The number of effective particles remains close to the total number of particles in all the experiments.
\end{abstract}
 
Several applications which range from  meteorology, oceanography, and hydrology to biological systems need to estimate the state of a system dealing  with partial noisy observations.  In these cases, the existence of physical constraints or laws that govern the evolution of the system, in general referred as dynamical models, is essential even if they are an imperfect or coarse description of the system evolution. The combination of a dynamical model which represents the nonlinear evolution of the (hidden) state  with sparse noisy observations is usually represented through a hidden Markov model, also known as state-space model \citep{doucet00,cappe05}. The aim of this work is to develop a methodology for sequential Bayesian inference of state-space models composed \pj{of} nonlinear chaotic dynamical systems in high-dimensional state spaces \pj{with non-Gaussian uncertainty in the state-space variables}.

Particle filters are Monte Carlo techniques that sample the space through so-called particles--state realizations-- \pj{that} represent the probability density of the state conditioned on the partial noisy observations, i.e. the posterior density. The challenge for particle filters is to represent recursively the high probability regions of the posterior density with a limited number of particles. In general, particle filters suffer from filter degeneracy \citep{doucet00}. After a few time iterations, they tend to give all the weight to a single particle. \linelabel{ln:proposal} \ma{In the importance sampling framework, a proposal distribution of the transition density conditioned to the current observations can be chosen to improve the sampling} \citep{doucet00,arulampalam02}. Although, a good choice of a proposal density may alleviate the filter degeneracy, if the filter is applied recursively a resampling step is still required. However, resampling produces another issue, particle impoverishment, \ma{since only a few particles with large weights are kept and copied.} This is an important contention point for hidden-Markov models \ma{in} high-dimensional state spaces.

Chorin et al., \cite{chorin09}, proposed an implicit sampling \pj{scheme} in which the particles from \pj{a Gaussian} proposal density are mapped to the high probability regions of the posterior density through the solution of an algebraic equation for each particle. The last relates the mode of the proposal density with the mode of the posterior density for each particle. To find the mode of the posterior density for each particle, a minimization is required.
Van Leeuwen, \cite{vanleeuwen10}, \pj{used a relaxation towards observations between observation times as proposal density.} To further \pj{reduce} the variance of the weights, that work also proposes a mapping that enforces equal weights by scaling the deterministic movement of particles in the optimal proposal step. \linelabel{ln:overly} \ma{Since the high probability region of the proposal density is chosen smaller than the} \pj{model transition density}, the filter can be overly optimistic about its performance \citep{vanleeuwen15}. Reich, \cite{reich13}, applied optimal transport principles to filters. \ma{A linear ensemble transform is proposed that minimizes the Euclidian distance with optimal coupling}. The degeneracy problem is still present in this method. It is, however, well suited for localization procedures for high-dimensional applications \citep{chen15}\pj{, in which an observation only affects the region of the state space that is close to it in Euclidian distance.}

In this work, we propose a novel particle filter which is also based on a mapping from the prior density to the posterior density as the implicit sampling filter \citep{chorin09,atkins13}, the implicit equal weight particle filter \citep{vanleeuwen10,zhu16} and the nonparametric ensemble transform method \citep{reich13}. \pj{The main difference is that the mapping is done iteratively }~\ma{via a gradient flow.} A sampling methodology based on optimal transport is derived to drive the particles from the proposal density to the posterior density. The transport of particles aims to minimize the Kullback-Leibler divergence that measures the differences between the \pj{the intermediate} densities and the posterior density. The mapping is embedded in a reproducing kernel Hilbert space which allows to obtain \pj{an analytical expression for} the gradient of the Kullback-Leibler divergence. 
The proposed sampling method for the particle filter is based in the Stein variational gradient descent introduced in \cite{liu16}. 
They found a connection between the gradient Kullback-Leibler divergence and the Stein discrepancy. In this work, we develop a sequential Monte Carlo filter based on this variational mapping, evaluate it in three hidden Markov models and discuss its usefulness for high-dimensional applications.

\section{Methodology}
\subsection{Sequential Bayesian estimation}\label{sec:HMM}

We assume the estimation problem is encompassed of a dynamical model $\dynm$, which predicts the state $\v x$ from a previous state, and an observational model $\obsm$, which transforms the state from the (hidden) state space to the observational space. The \pj{set of} equations that \pj{defines} the estimation problem is known as a state-space model or a hidden Markov model. These are
\mi
\v x_k = \dynm(\v x_{k-1},\gv \eta_{k}), \label{HMMdyn}
\mf
\mi
\v y_k = \obsm(\v x_{k},\gv \nu_{k}), \label{HMMobs}
\mf
where $\v x_k\in \mathbb{R}^{N_x}$ is the state at time $k$, $\v y_k\in \mathbb{R}^{N_y}$ are the observations, \linelabel{ln-peta} $\gv \eta_{k}\sim p(\gv \eta)$ is the random model error, and $\gv \nu_{k}\sim p(\gv \nu)$ is the observational error.
\ma{The method developed here is general and does not rely on the additive or Gaussian assumption in model errors.}

Observations $\v y_k$ are assumed to be measured at \pj{discrete times}. Some apriori knowledge of the state at $k=0$ is assumed, the initial prior state density $p(\v x_0)$. The sequential Bayesian state inference is given in two stages:

\begin{enumerate}
\item Firstly, in the  evolution stage the prediction density is determined as
\mi
p(\v x_k|\v y_{1:k-1})=\int p(\v x_k|\v x_{k-1}) p(\v x_{k-1}|\v y_{1:k-1}) \, \ud \v x_{k-1}, \label{forcstPDF}
\mf
where we denote $\{\v y_1,\cdots,\v y_{k-1}\}$ as $\v y_{1:k-1}$. At $k=1$, $\v y_{1:0}=\varnothing$ so that the prediction density is  $p(\v x_1)=\int p(\v x_1|\v x_{0}) p(\v x_{0}) \ud \v x_{0}$.

\item Secondly, in the assimilation stage Bayes rule is used to express the inference as a sequential process
\mi 
p(\v x_k|\v y_{1:k})=\frac{p(\v y_k|\v x_k) p(\v x_k|\v y_{1:k-1})}{p(\v y_k|\v y_{1:k-1})} \label{posteriorPDF}
\mf
where $p(\v y_k|\v x_k)$
is the observation likelihood and $p(\v y_k|\v y_{1:k-1})=\int p(\v y_k| \v x_k) p(\v x_k|\v y_{1:k-1}) \ud \v x_k$ is the marginalized likelihood.

\end{enumerate}

\subsection{Particle flows and optimal transport}\label{sec:flows}

\ma{The concept of homotopy can be used to transform the \pj{prior} density towards the \pj{posterior} density. Continuous deformations through a parameter \pj{can be used to achieve this, such as} $T(\v x,\lambda): \mathbb{R}^{N_x}\times [0,1]\rightarrow \mathbb{R}$, $T(\v x,\lambda)=p(\v x)^\lambda q(\v x)^{1-\lambda}$. (\pj{To simplify the notation  we have left the conditioning to observations and time index implicit in this and the following subsection.}) The parameter $\lambda$ is interpreted as a pseudo-time which is varied from 0 to 1 at a fixed real time. This concept was used in \cite{daum07} for particle filters.  The densities are represented through particles, so that the deformations are represented by particles moving in a flow according to a set of ordinary differential equations,}
\mi
\frac{d \v x_\lambda}{d \lambda}=\v v_\lambda(\v x_\lambda), \label{flow}
\mf
\ma{where $\v v_\lambda$ is the drift or velocity field. We focus on deterministic flows so that diffusivity processes are not considered. In some recent works, the particles are pushed smoothly forward in pseudo-times using a Gaussian approximation to represent the velocity field  (e.g. \cite{bunch16,li17}). }

\ma{Another promising approach for the sampling of complex posterior distributions is the optimal transport problem \citep{marzouk17}. Given  the \pj{prior probability} mass distribution $q(\v x)$ and  the target \pj{probability} mass distribution $p(\v z)$, we want to transport the \pj{probability mass} from $q$ to $p$ using a mapping $T: \v x\rightarrow \v z $. The optimal transport problem seeks the transformation $T$ that gives the minimal cost to transport the distribution of mass from $q$ to $p$.  This represents the classical Monge optimal transport problem. There is a rigorous proof of the existence of such optimal mapping under mild conditions \citep{mccann95}. Furthermore, \citep{angenent03} have formulated the Monge-Kantorovich problem as a steepest descent gradient optimization. }

\ma{Our approach combines the ideas of optimal transport and particle flows, the so-called local approach to optimal transport \citep{villani08}. Through a sequence of mappings we seek to push forward the particles from the \ma{prior} to the target density. As in particle flows, the sequence of mappings are required to be as smooth as possible. The particles behave as active Lagrangian tracers in a flow. The velocity field at each pseudo-time step of the mapping sequence is chosen following a local optimal transport criterion. 
}

\subsection{Variational mappings}\label{sec:var}

At each iteration of the sequence, we propose a local transformation $T$ that follows \reff{flow},
\mi
\v x_{\lambda+\epsilon}= T(\v x_\lambda)= \v x_\lambda + \epsilon \v v(\v x_\lambda), \label{transforT}
\mf
where $\epsilon=\delta \lambda$ is  assumed to be small and $\v v(\v x_\lambda)$ is an arbitrary vectorial function ``sufficiently'' smooth, which represents the velocity of the flow defined in \reff{flow}.

The Kullback-Leibler  divergence is used as a measure of the difference between the \pj{intermediate density $q(\v x)$ and the posterior density $p(\v x)$.} The Kullback-Leibler  divergence between $q$ and $p$ after the transformation, $T$, in terms of the inverse mapping is
\mi
KL_T = \int q_X(\v x) \log\left(\frac{q_X(\v x)}{p_{T^{-1}}(\v x)}\right) \ud \v x, \label{KL_inv}
\mf
where $p_{T^{-1}}(\v x)=p(T(\v x)) \det{\nabla_x T(\v x)}$ and $\det{\nabla_x T(\v x)}$ is the determinant Jacobian of the mapping  and for brevity $\v x=\v x_\lambda$.

Our goal is to find the velocity field $\v v(\v x)$ of the transformation $T$  that gives the minimum of $KL_T$ for each mapping of the sequence. In other words, we need to find the direction $\v v (\v x)$, that gives the steepest descent of $KL$.

We use the Gateaux derivative of a functional, which is a generalization of the directional derivative. Given the functional $F$, it is defined as
\mi
D_h F = \lim_{\epsilon\rightarrow 0 }  \frac{F(\v x+\epsilon \v h(\v x))-F(\v x)}{\epsilon}= \ud_\epsilon F(\v x+\epsilon \v h)|_{\epsilon=0}.
 \mf
  
The Gateaux derivative of the Kullback-Leibler  divergence, \reff{KL_inv}, in the direction $\v v(\v x)$ is given by
\mi
D_v KL = - \int q(\v x) \left. \ud_\epsilon \log  p_{T^{-1}}(\v x)\right|_{\epsilon=0} ~\ud \v x . \label{DKL1}
\mf
The derivative of the transformed log-posterior density is 
\mi
\ud_\epsilon \log p_{T^{-1}}(\v x) = \nabla_T \log p(T(\v x))^\transp \, \ud_\epsilon T + \mathrm{Tr}(\nabla_x T^{-1} \ud_\epsilon\nabla_x  T)|_{\epsilon=0}. \label{dlogp_epsilon}
\mf
Considering that $T=\v x + \epsilon \v v$ in \reff{dlogp_epsilon} and replacing in \reff{DKL1}, the directional derivative of $KL$ results in
\mi
D_v KL = - \int q(\v x) \left[ \nabla_x \log p(\v x)^\transp \v v(\v x) + \mathrm{Tr}(\nabla_x \v v) \right] ~\ud \v x. \label{dKL_dphi}
\mf

Equation {\bf\ref{dKL_dphi}} gives the $KL$ derivative along $\v v$. However, we require the negative gradient of $KL$ \ma{in terms of the samples of $q$} for the optimization of $KL$ as a function of $T$. In general $\v v$ belongs to an infinite dimensional Hilbert space.  Hence, the full optimization problem is still intractable in practice since we do not have a way to determine the $-D_{v} KL$  that gives the steepest descent direction.

One way to limit our functional optimization problem so that it becomes tractable is choosing as space of functions the unit ball of a reproducing kernel Hilbert space (RKHS), which we denote as $\rkhs$. This was proposed by \cite{liu16}. In this way, we constrain $\v v$ to  $\rkhs$ and  require that $\parallel \! \v v \!\parallel_{\rkhs}\le 1$  to find the gradient of $KL$. The optimization problem is then to find the $\v v \in \rkhs$ that gives the direction of steepest descent of $KL$. The main properties of the RKHS are included in the SI Appendix.

Given a vector-valued kernel function, $\v K = K \v I$, any function from $\rkhs$ can be expressed as the dot product by the kernel that defines the RKHS,
\mi
\v v(\v x) =\langle \v K(\cdot , \v x), \v v(\cdot)\rangle_{\rkhs}. \label{dotK-phi}
\mf
Equation \ref{dotK-phi}  is known as the reproducing property (see SI Appendix SI.B). The scalar kernel $K$ defines  $\rkhs^1$.

Using \reff{dotK-phi} in \reff{dKL_dphi} and dot product properties, \pj{we find:}
\mi
D_v KL = \langle - \int q(\v x) \left[ K(\v x, \cdot ) \nabla_x \log p(\v x) +  \nabla_x K(\v x, \cdot )\right] ~\ud \v x, \,\, \v v(\cdot)\,\rangle_{\rkhs^1}. \label{DKL_final}
\mf
This is valid for any $\v v$ such that $\parallel \! \v v \! \parallel_\rkhs \le 1$. Therefore, the first term of the dot product in \reff{DKL_final} is by definition the gradient of $KL$ at $T_{\epsilon=0}=\v x$,
\mi
\nabla KL(\v x) = - \mean_{x'\sim q}\left[ K(\v x', \v x ) \nabla_x \log p(\v x') +  \nabla_x K(\v x', \v x)\right]. \label{KLgrad1}
\mf
\ma{This expression for the gradient, \reff{KLgrad1}, is particularly suitable for Monte-Carlo integration when $q$ is only known through a set of samples.} Since we seek to minimize the Kullback-Leibler  divergence we choose as direction of the transformation $T$ the negative of its gradient, the steepest descent direction,
\mi
\v x_{\lambda+\epsilon} = \v x - \epsilon \nabla KL(\v x). \label{graDesc}
\mf

\linelabel{ln:gradientflow} Using Kernel reproducing property and integration by parts in \reff{KLgrad1}, the resulting gradient flow of the mapping is
\mi
\v v_\lambda(\v x)= q_\lambda(\v x) \nabla \log p(\v x) - \nabla q_\lambda(\v x). \label{optimalflow}
\mf
\linelabel{ln:convergence}
\ma{In \cite{moselhy12},  the  Kullback-Leibler  divergence is also used \pj{but with an extra} regularization term to find a single global optimal map. Here, we apply a gradient flow via a sequence of local mappings. Under smoothness constrains\citep{villani08}, the minimum of the cost function as a function of the mappings, \reff{transforT}, is uniquely determined in \reff{graDesc}, i.e. no regularization term is required. As shown in \cite{tabak10}, the gradient flow, \reff{optimalflow}, has as stationary solution $q(\v x)=p(\v x)$ so that the gradient flow converges toward the target density.}

\subsection{Mapping particle filter}\label{sec:mpf}

Consider we have a set of \pj{equal-weight} particles  $\v x_{k-1}^{(1:N_p)}$  
which sample the posterior density at time $k-1$. The target density at time $k$ which we aim to sample using the variational mapping is the posterior density $p(\v x_k|\v y_{1:k})$. \linelabel{rev-method} \ma{The mapping approach only requires a set of samples of the \pj{prior} density. It is started by the set of unweighted particles that are evolved from the previous estimate to the \pj{present} assimilation time by the dynamical model, i.e. $\left\{\v x_{k,0}^{(j)}=\dynm(\v x^{(j)}_{k-1})+\gv \eta^{(j)}_k\right\}_{j=1}^{N_p}$, where the second subscript represents the mapping iteration. These samples are then pushed towards the sequential posterior density by the mapping iterations.} 

Given the set of particles $\v x_{k,i-1}^{(1:N_p)}$ that are samples of the \pj{intermediate} density at mapping iteration $i-1$, the gradient of the Kullback-Leibler  divergence from \reff{KLgrad1}  at a state space position $\v x$ by Monte-Carlo integration is 
\begin{align}
\nabla KL(\v x) = - \frac{1}{N_p} \sum_{l=1}^{N_p} & \left[ K(\v x_{k,i-1}^{(l)}, \v x ) \nabla \log p(\v x_{k,i-1}^{(l)}) \right. \nonumber \\
    & \left.  +  \nabla_x K(\v x_{k,i-1}^{(l)}, \v x)\right].\label{gradKL1}
\end{align}

The negative of this gradient represents the velocity field of the gradient flow. An interpretation of the two RHS terms in \reff{gradKL1} is given in SI Appendix (SI.C).
At the mapping iteration $i$, the particle $j$ is transformed, according to the mapping \reff{graDesc}, by
\mi
\v x^{(j)}_{k,i}=\v x^{(j)}_{k,i-1} - \epsilon \, \nabla KL (\v x^{(j)}_{k,i-1}), \label{optim}
\mf
where $-\nabla KL  (\v x^{(j)}_{k,i-1})$ is the steepest descent direction at the particle position. \reff{optim} may be interpreted as the movement of the particles along the streamlines of the flow assuming small $\epsilon$.

 \pj{An ingredient needed to evaluate \reff{gradKL1} is an expression for the gradient of the posterior density.
A problem in sequential Bayesian inference is that there is no exact expression for the posterior density. We do know the likelihood function, 
but we only have a set of particles that represent the prior density, not the density itself.} The prior density using the particle representation of the posterior density at time $k-1$ in \reff{forcstPDF} results in
\mi
p(\v x_k|\v y_{1:k-1})  \approx  \frac{1}{N_p} \sum_j^{N_p} p(\v x_k | \v x^{(j)}_{k-1}),
\mf
and the expression for the posterior density from \reff{posteriorPDF} becomes
\mi
p(\v x_k | \v y_{1:k})  \propto  \frac{1}{N_p} p(\v y_k | \v x_k) \sum_j^{N_p} p(\v x_k | \v x^{(j)}_{k-1}). \label{targetden}
\mf
    
Because of the finite ensemble size at time $k-1$, this expression is an approximation of the posterior density. Equation \ref{targetden} is the target density in the mapping.

\ma{At each mapping iteration, all the particles are moved following \reff{optim}. Successive applications of the transformation, through gradient descent, with the corresponding updates of $\nabla KL$, \pj{will} converge towards the minimum of the Kullback-Leibler divergence. The pseudo-code of the implemented algorithm which combines variational mapping with the sequential particle filter is shown in Algorithm~\ref{algo}.}

 \ma{The form in which the variational mapping is obtained, as steepest gradient descent, is suitable for the stochastic optimization algorithms used in the machine learning community.\linelabel{adaptative-method} In this case, $\epsilon$, known as the learning rate, can be determined adaptively using $\nabla KL$ from previous iterations (e.g. \cite{zeiler12}).}

\begin{algorithm}
\caption{Mapping particle filter algorithm}\label{algo}
\small
\begin{algorithmic}
\Require{Given $\v x_{k-1}^{(1:N_p)}$,  $\v y_k$, $\mathcal{M}(\cdot)$, $p(\v y|\v x)$ and $p(\gv\eta)$}
  \For{$j=1,N_p$}
    \Let{$\v x_{k,0}^{(j)}$}{$\mathcal{M}(\v x_{k-1}^{(j)})+\gv \eta_{k}$}\Comment{Forecast stage}
    \EndFor

\Repeat  \Comment{Mapping iterations}
   \For{$j=1,N_p$}
   \Let{$\v x^{(j)}_{k,i}$}{$\v x^{(j)}_{k,i-1}- \epsilon \, \nabla KL(\v x^{(j)}_{k,i-1})$}\Comment{$\nabla KL$ from \reff{gradKL1}}
   \EndFor
   \Let{i}{i+1}
\Until Stopping criterion met
\Ensure{$\v x_{k,i}^{(1:N_p)}$}
\normalsize
\end{algorithmic}
\end{algorithm}

\pj{An important part of any iterative variational method is a stopping criterion. One possibility is to check the value of $| \nabla KL|$
for each particle, or averaged over all particles. Another option is to use importance sampling and to interpret the final density of the mapping sequence as a proposal density
and calculate weights with respect to the posterior density, which will then automatically lead to an unbiased estimator.
To use this, an expression for the final proposal density $q$ is required, however, we have only its particle representation.
Two possibilities exist to implement this in a practical way. If the dimension of the system is small one can use kernel-density
estimation based on the particles. The other approach is to use the transformations, that we do have explicitly,
and relate the final density to the prior density. The approaches are detailed in the SI Appendix (SI.D).}



\ma{To avoid the resampling in the importance sampling step, one can calculate the effective ensemble size as $N_{eff}$, and 
decide to continue the iterations in Algorithm~\ref{algo} until $N_{eff}$ reaches a certain threshold $N_t$. In the following assimilation cycle, the weights of the particles representing the final proposal density has to be considered in the sequential posterior density, resulting in a weighted posterior density instead of \reff{targetden}. One should keep in mind, however, that our estimate of the posterior density is rather poor in high-dimensional systems with a small number of particles, so the weights would not be very accurate. }

\section{Numerical experiments}\label{sec:nume}

The implementation of the mapping particle filter is evaluated using three chaotic nonlinear dynamical models with state spaces of up to 40 dimensions. \pj{Higher dimensional state spaces $>10^3$ will be pursued in a followup work.} We conducted a set of stochastic twin experiments. In these experiments, the dynamical model to produce the synthetic observations is the same as the one used in the state-space model, including the statistical parameters.


A Gaussian kernel is chosen for the experiments,
\mi
K(\v x,\v x')=\exp\left[-\frac{1}{2}(\v x-\v x')^\transp \v A^{-1} (\v x-\v x')\right].
\mf
As a first reasonable choice, $\v A$ is taken proportional to the model error covariance matrix,  $\v Q$, i.e. $\v A=\alpha \v Q$. \linelabel{matrixA}\ma{This appears to be a convenient choice since the model uncertainty, $\v Q$, already includes the physics so that it is expected to represent the scaling between the variables and also the correlations between variables. Under this choice the only parameter of the kernel that requires to be defined is $\alpha$.} \pj{With growing dimension of the state vector and the number of particles small, the distance between the particles grows larger. To account for this, $\alpha$ should be chosen larger to increase the distance between particles. In the experiments, we have found that $\alpha$ should be chosen of the order of the dimension of the state vector}. Further details on the experiments are given in SI Appendix.

\begin{figure}
  \hskip -0.3cm
  \centering \includegraphics[width=4.2in]{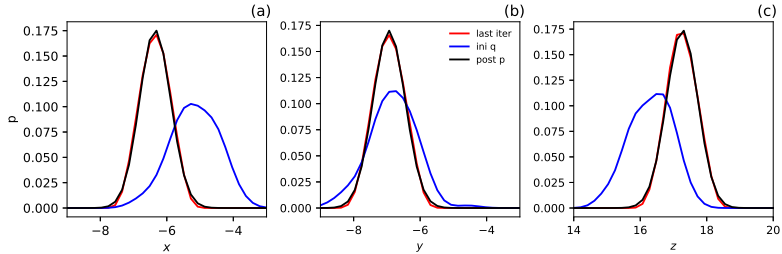}

\vskip -.3cm
\caption{Marginalized prior density determined at the 115-th time iteration, $k=115$ (blue line), the final density after the variational mapping (red line) and the marginalized sequential posterior ``target'' distribution (black). }\label{post1}
\end{figure}


\begin{figure}
\hskip -0.3cm
\centering\includegraphics[width=4.2in]{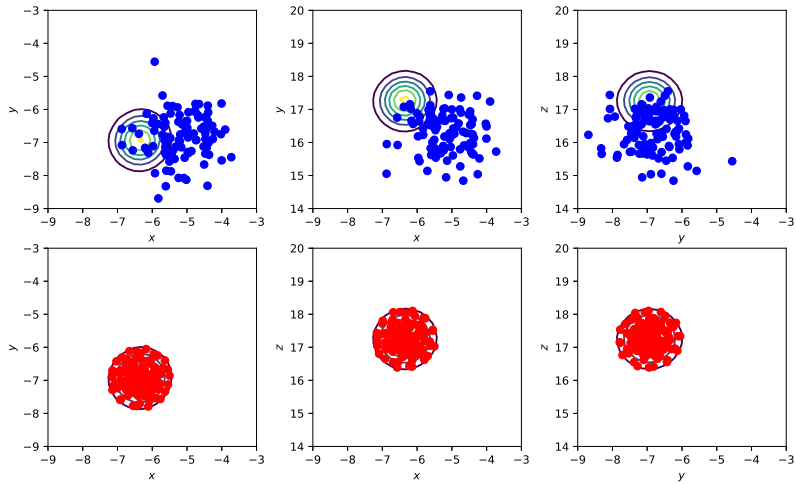}
\vskip -.3cm
\caption{Marginalized posterior density (contours) and the particles sampling the prior density at the 115-th time iteration for the three sections at the mode location (upper panels).  The particles sampling the final density after the variational mapping with 50 iterations (lower panels). }\label{post2}
\end{figure}

A quick convergence of the \ma{prior} to the posterior density in terms of the mapping iterations  is the key ingredient for high-dimensional applications. The applications we have in mind satisfy $N_p\ll N_y < N_x$. Therefore, we evaluated, apart from stochastic gradient descent with a fixed learning rate $\epsilon$,  adadelta \citep{zeiler12} and adam \citep{kingma15} which have an adaptive learning rate for each state variable (see SI Appendix, SI.F). These stochastic optimization methods are particularly suitable for high-dimensional spaces.

The maximum number of mapping iterations is fixed to $I=50$, and a threshold for the number of effective particles is used as criterion of convergence, $N_t=0.9 N_p$.  As mentioned,  a threshold in the absolute value of the gradient could also be used as criterion of convergence.  In SI Appendix, SI.F we compare these two criteria for convergence of the variational mapping.

\section{Results}\label{sec:results}

Figure \ref{post1} shows the marginalized prior density as a function of the three state variables of the Lorenz-63 system determined at the 115-th time assimilation cycle, $k=115$ (each panel exhibits a  variable). \ma{For plotting, the marginalized density is determined with kernel density estimation.} The cycle was chosen so that the differences between the  mapping particle filter (MPF) and the sampling importance resampling (SIR) filter are emphasized (i.e. a cycle for which the apriori density is not a good representation of the posterior density). The experiment uses 100 particles and the full state is observed. Because particles are spread in and out the high observation likelihood region, the proposal density is broader, asymmetric compared with the marginalized sequential posterior density. 
After the variational mapping, the resulting density is a close representation of the sequential posterior density, the target density for the mapping, in the three variables (see the three panels in Fig. \ref{post1}).

The distribution of the particles can be seen in Fig. \ref{post2}. Upper panels of Fig. \ref{post2} show the particles belonging to the \pj{prior density for each variable, the initial density for the mapping.} Contours show the posterior density, \pj{marginalised to the 2-dimensional plane}. Lower panels show the final locations of the samples after 50 mapping iterations. The mapping has pushed most  particles towards the high probability region of the posterior density. Note that \pj{the} algorithm is not collapsing the particles toward the mode of the posterior, but representing the density with the samples as a whole. The diversity of the particles produced by the mapping is notable.

\begin{figure}
\hskip -0.3cm
\centering\includegraphics[width=4.2in]{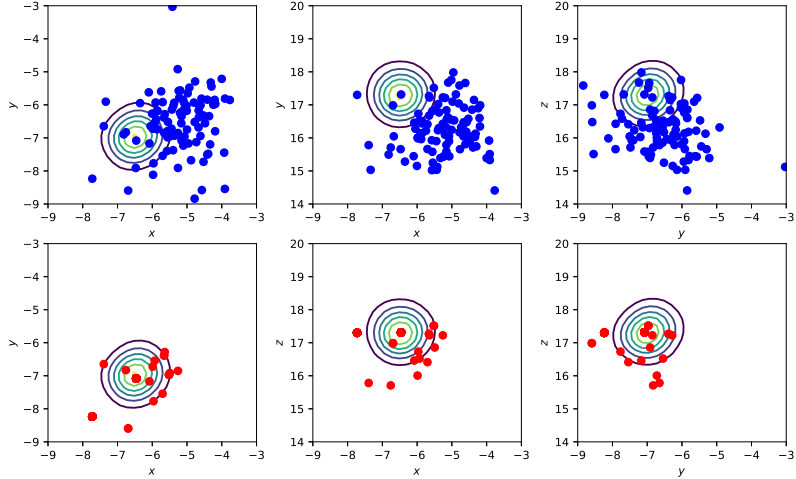}
\vskip -.3cm
\caption{As in Fig.~\ref{post2} for the SIR particle filter. }\label{postsir2}
\end{figure}

The same experiment  conducted  with the standard SIR\pj{, or bootstrap} filter using resampling when $N_{eff}<N_p/2$ is shown in Fig. \ref{postsir2}. 
The sampling deficiencies in the SIR filter are visible in Fig. \ref{postsir2} where the particles of the \pj{prior} density (upper panels) and the resampled density (lower panels) are shown with dots. Because the resampling conserves and replicates statistically only the particles with high likelihood, there is an evident sample impoverishment even in this low-dimensional twin experiment.

\begin{figure}
\hskip -0.3cm
\centering\includegraphics[width=4.2in]{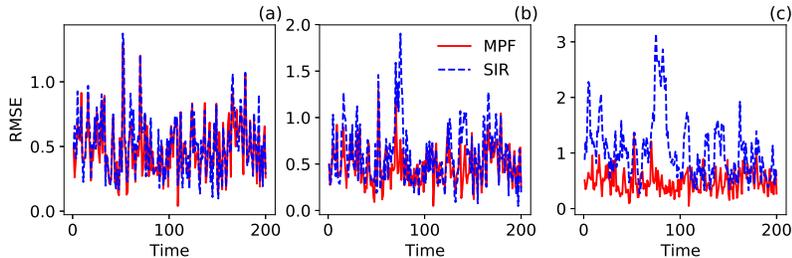}
\vskip -0.3cm
\caption{RMSE for the MPF and the SIR filters as a function of the number of particles. Panels (a) 100, (b) 20, and (c) 5 particles.}\label{rmse}
\end{figure}

\ma{Because of the good sampling, the MPF exhibits a very weak sensitivity to  the number of particles in the root-mean-square error (RMSE) measure}. Figure \ref{rmse} shows the RMSE of the analysis mean with respect to the true state, for  MPF and SIR filters for 100, 20 and 5 particles (Fig. \ref{rmse} a, b and c respectively). Even for a large number of particles (with respect to the state dimension), the MPF outperforms the mean estimation with respect to the SIR filter. SIR filter diverges for 5 particles. Figure \ref{rmse} a, b and c shows that the time-mean RMSE for the MPF practically does not change between 100 particles with a time-mean RMSE of 0.482 to 5 particles whose time-mean RMSE is 0.489. Thus, the performance of the MPF is quite robust, only small changes in the RMSE are found when decreasing the number of particles. 

\linelabel{complexity}\ma{Note the complexity of Algorithm~\ref{algo} is proportional to $N_p^2$ while \pj{it is proportional to $N_p$ for the SIR filter}, so that the computational cost of the assimilation stage is \pj{higher} for the MPF with the same \pj{number of} particles. However, a smaller number of particles is required so that it represents a promising venue for expensive dynamical models.}

\begin{figure}
\begin{center}
\centering\includegraphics[width=3.4in]{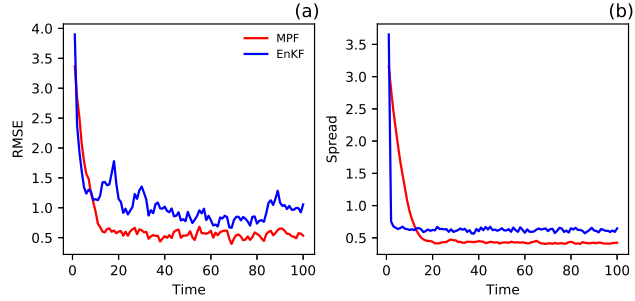} 
\end{center}
\vskip -4mm
\caption{RMSE (a) and spread (b) of the state as a function of the assimilation cycles produced by  EnKF and MPF  in the 40-variables Lorenz-96 system.}\label{figL96}
\end{figure}

To evaluate the performance of the MPF for a larger state-space model, an experiment using the 40-variable Lorenz-96 dynamical system was conducted. An ensemble of 20 particles was used in the experiment. \linelabel{ln:scalable} \ma{Scalable sampling methods are expected to produce useful results for these experiments in which $N_p<N_x$.} Figure \ref{figL96}a compares the resulting RMSE of the state given by MPF  and \ma{the \pj{stochastic} EnKF without localization as a function of time. \pj{The SIR filter is not shown since it} exhibits degeneracy in this experiment because of the high dimensional state and observation spaces.} 
The MPF takes a longer time to converge (20 cycles), but the \ma{EnKF exhibits a larger RMSE (this result might be improved using more members or localization,} \pj{although the Gaussian assumption will hit the EnKF results)}. \pj{Furthermore, the mapping particle filter estimates are rather stable in time, which is a result of its deterministic mapping in which the particles feel each other.} The spread is rather stable and comparable to the RMSE in both filters (Fig. \ref{figL96}b).

\begin{figure}
\vskip -.4cm
\hskip -0.3cm
\centering\includegraphics[width=4.2in]{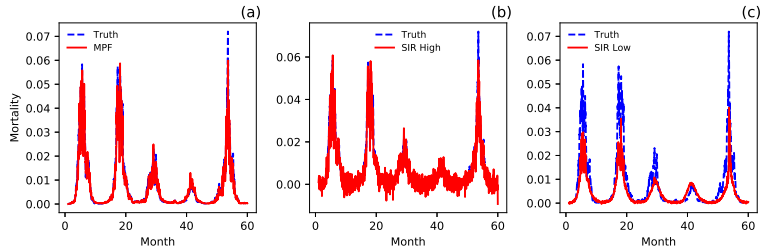}
\vskip -.3cm
\caption{Mortality time series for a model of cholera dynamics, true mortality and the one estimated with the MPF (a), with the SIR filter with high additive noise (b) and with the SIR filter with low additive noise. }\label{figcholera}
\end{figure}

A third set of experiments was conducted using a compartment stochastic dynamical model for cholera dynamics \citep{ionides06}. This is a 5-variable system allowing for susceptible, infected, and three classes of recovery individuals in which cholera mortality is the only observed variable. \linelabel{ln:challenges} From the inference point of view, it presents some challenges\citep{ionides06}.  It has partial noisy observations whose variance depends on time. Transmission is assumed stochastic and \pj{has a multiplicative model error}. Since \pj{our implementation of }the MPF assumes additive model error, we have augmented the state space with a new variable \pj{with an evolution equation equal to the noise factor} and in this way the augmented cholera model can be represented by additive model error terms. The ensemble is composed of 100 particles and a small artificial additive model error term was added to benefit particle diversity \citep{liu01}. \ma{Note that a non-Gaussian density for \pj{additive} model error would give more realistic features.} Figure \ref{figcholera}a shows the true mortality time series and the one estimated by the MPF. \pj{The experiment started with an initial mean state which underestimates the number of true susceptible individuals by 20\%} . A good performance is found for the MPF \ma{(total RMSE is 0.0031)}. \ma{The SIR filter follows the cholera outbreaks but the estimate is much more noisier when the same level of noise is used as for the MPF (total RMSE is 0.0039). By decreasing the level of noise in the SIR filter, the noise diminishes but at the expense of a strong underestimation of the peaks (total RMSE is 0.0077). }

\section{Discussion}\label{sec:conclu}
The proposed variational mapping particle filter is based on a deterministic gradient flow. \linelabel{ln:resamplingconv} It is able to keep a large effective sample size, avoiding the resampling step, even for long recursions \ma{subject to the convergence of the mapping sequences}. Furthermore, it shows a robust behavior in terms of the RMSE \pj{with a consistent ensemble spread \ma{in the conducted experiments}, using only a small number of particles.} 

\linelabel{3dvar} In the limit of a single particle\ma{, $N_p=1$, the gradient of the Kullback-Leibler divergence is equal to the negative of the gradient of the log-posterior density (see \reff{gradKL1}).  In that case, the optimization via gradient descent determines the mode of the posterior density.} Therefore, the method for $N_p=1$ is equivalent to three-dimensional variational data assimilation (3D-Var) with $\v Q$ in the role of $\v B$. 

The mapping of samples in the variational mapping particle filter is produced by moving the particles where they maximize the information gain of the proposal density with respect to the sequential posterior density, via the Kullback-Leibler divergence.  In these terms, the proposed mapping particle filter seeks to maximize  the amount of information available in a complex high-dimensional posterior density \pj{using a stochastic optimization method and given a limited number of particles.}

{\bf Acknowledgements.} We are grateful to the reviewers for valuable comments. This work has been founded by CUNDA project of the European Research Council. A python library with the mapping particle filter is available at \url{http://github.com/pulidom/mpf/}.

\clearpage

\renewcommand{\thefigure}{S\arabic{figure}}
\setcounter{figure}{0}
\renewcommand{\theequation}{S\arabic{equation}}
\setcounter{equation}{0}
\renewcommand\thesection{SI}
\renewcommand\thesubsection{SI.\Alph{subsection}}

\section{Supporting Information}\label{SI}

\subsection{Hidden Markov models with Gaussian errors}

Next, we derive the expression of the posterior density in the sequential framework as a function of the particles at $k-1$. For the numerical experiments, model and observational errors are assumed Gaussian. These are taken as an example, but the technique framework is general. In particular, it is suitable for non-additive and non-Gaussian errors. The prediction probability density can be obtained analytically under the additive Gaussian model error assumption, $\gv \eta_k \sim \gaus(\v 0, \v Q_k)$ in \reff{HMMdyn}. The evolution of the particles from  \reff{forcstPDF} gives
\mi
p(\v x_k|\v y_{1:k-1}) \propto \sum_{m=1}^{N_p}  \exp \left[-\frac{1}{2}\lVert\v x_k-\dynm(\v x_{k-1}^{(m)})\rVert^2_{\v Q_k}\right],\label{MCforecastPDF}
\mf
where $\norm{\v x_k-\dynm(\v x_{k-1}^{(m)})}^2_{\v Q_k}\doteq (\v x_k-\dynm(\v x_{k-1}^{(m)}))^\transp \v Q_k^{-1} (\v x_k-\dynm(\v x_{k-1}^{(m)}))$. \pj{We assume equal-weighted particles here; it is straightforward to assume weighted particles at time $k-1$ if so desired.}

The observational error is also assumed \pj{additive} Gaussian, $\gv \nu_k \sim \gaus(\v 0, \v R_k)$ in \reff{HMMobs}, so that the observational likelihood at $k$ is
\mi
p(\v y_{k}|\v x_k) \propto  \exp \left[-\frac{1}{2}\norm{\v y_k- \obsm(\v x_k)}_{\v R_k}^2\right].\label{GaussianObsLik}
\mf

The Bayes rule, \reff{posteriorPDF}, is used to obtain the sequential posterior density combining the prediction density, \reff{MCforecastPDF}, and the observation likelihood, \reff{GaussianObsLik}. The sequential posterior density given the particles at the previous time step is proportional to
\begin{multline}
p(\v x_k|\v y_{1:k}) \propto  \sum_{m=1}^{N_p}  \exp \left[-\frac{1}{2}\norm{\v x_k-\dynm(\v x_{k-1}^{(m)})}_{\v Q_k}^2 \right]\\
 \cdot \exp \left[-\frac{1}{2}\norm{\v y_k- \obsm(\v x_k)}_{\v R_k}^2 \right] . \label{anaPost}
\end{multline}

This is the target density that the particles $\v x_k^{(1:N)}$ as Monte Carlo samples should be representing in the filter.

The gradient of the logarithm of the sequential posterior density, \reff{anaPost}, at $\v x^{(l)}_{k,i-1}$ is given by
\begin{multline}
  \nabla_{\v x} \log p(\v x^{(l)}_{k,i-1}) = C p^{-1}(\v x^{(l)}_{k,i-1}) \sum_{m=1}^{N_p}  \psi_{l,m}^Q \psi_{l}^R\\
  \left\{\v H^\transp \v R_k^{-1} \left[\v y_k- \obsm(\v x^{(l)}_{k,i-1})\right]- \v Q_k^{-1} \left[\v x^{(l)}_{k,i-1} - \mathcal{M}(\v x^{(m)}_{k-1})\right]\right\}\label{gradp-1}
\end{multline}
with $\psi_{l,m}^Q=\exp\left[-\frac{1}{2} \norm{x_{k,i-1}^{(l)}-\mathcal{M}(\v x^{(m)}_{k-1})}_{\v Q_k}^2\right]$,\\
$\psi_{l}^R= \exp \left[-\frac{1}{2}\norm{\v y_k- \obsm(\v x_k^l)}_{\v R_k}^2 \right]$ and $p(\v x^{(l)}_{k,i-1})= C \sum_{m=1}^{N_p} \psi_{l,m}^Q \psi_{l}^R$. Reducing we obtain
\begin{multline}
\nabla_{\v x} \log p(\v x^{(l)}_{k,i-1})=\v H^\transp \v R_k^{-1} \left(\v y_k- \obsm(\v x^{(l)}_{k,i-1})\right) \\
- \v Q_k^{-1} \left[\v x^{(l)}_{k,i-1} -\frac{\sum_{m=1}^{N_p}  \psi_{l,m}^Q \mathcal{M}(\v x^{(m)}_{k-1})}{\sum_{m=1}^{N_p}   \psi_{l,m}^Q}\right]. \label{gradp}
\end{multline}
This gradient of the log-posterior, \reff{gradp}, does not depend on the unknown normalization constant of the sequential posterior density. Hence, \ma{the numerical evaluation of the gradient of the Kullback-Leibler divergence is feasible.}

\subsection{Reproducing kernel Hilbert space}\label{sec:rkhs}

Kernel functions are used in machine learning for finding regressions in datasets using an implicit feature space \citep{scholkopf02}. Instead of determining the data representation in that space, only the inner products of the images of two data points in the feature space are required. Then, the inner products in the image space are replaced by kernels in the feature space, the so-called kernel trick. 


Some of the properties of the kernel $K$ are:

\begin{enumerate}
\item Let $K(\v x,\v x')$ be a positive definite kernel. Mercer's theorem says that the spectral decomposition
\mi
K(\v x,\v x')= \sum \lambda_j  e_j(\v x) e_j(\v x')
\mf
is given by a uniformly convergent series of the eigenfunctions  $e_j(\v x)$. The eigenfunctions satisfy $\langle e_i(\v x),e_j(\v x) \rangle_\rkhs= \delta_{ij}/\lambda_j$ and the eigenvalues are ordered, the larger the first.

\item Let $\rkhs^1$ be a RKHS, and $f$, $g$ two functions of $\rkhs^1$ such that $f(\v x)=\sum_j f_j  e_j(\v x)$, $g(\v x)=\sum_j g_j  e_j(\v x)$. The inner product of the functions in $\rkhs^1$ is given by
\mi
\langle f,  g \rangle_{\rkhs^1} = \sum_j f_j g_j/\lambda_j.
\mf
\item $\rkhs^1$ satisfies the reproducing property,
\mi
f(\v x)=\langle f,K(\cdot,\v x) \rangle_{\rkhs^1}, \qquad K(\v x,\v x')=\langle K(\v x,\cdot),K(\v x',\cdot)\rangle_{\rkhs^1}. \label{reproProp}
\mf
\item Every positive definite $K$ defines a unique RKHS.
\end{enumerate}
If $\v f=(\tilde f_1,\cdots,\tilde f_{N_x}) \in \rkhs$ where $\rkhs \doteq \rkhs^{N_x}$ and $\tilde f_j\in \rkhs^1$, then the inner product in $\rkhs$ is $\langle \v f,  \v g \rangle_{\rkhs}= \sum^{N_x}_{j=1} \langle \tilde f_j,  \tilde g_j \rangle_{\rkhs^1}$.

\subsection{Interpretation of the gradient of the Kullback-Leibler divergence}\label{interpreterm}

The first term of the gradient of the  Kullback-Leibler divergence, \reff{gradKL1}, tends to drive the particles towards the \pj{peaks}  of the posterior density. It gives a weighted average of $\nabla \log p$ including the current particle and surrounding ones within the kernel scale of the current one. This is the typical behavior of a variational importance sampler so that it accumulates particles at the high probability regions of the posterior density. On the other hand, the second term of $\nabla KL$ in \reff{gradKL1} tends to separate the particles. If the current particle $\v x_k^{(j)}$ is in the influence region of another particle say $\v x_k^{(l)}$, i.e. within the kernel scale, the term $\nabla_{\v x^{(l)}_{k,i-1}}  K(\v x^{(l)}_{k,i-1},\v x^{(j)}_{k,i-1})$ will tend to separate them acting as a repulsive force between particles.

As an example, suppose the radial basis functions as kernels, $K(\v x,\v x')=\exp\left(-\frac{1}{2 h} \parallel \v x -\v x' \parallel^2\right)$, the gradient of the kernel is 
\mi
\nabla K(\v x,\v x')= -\frac{1}{h} (\v x - \v x') K(\v x, \v x').
\mf

Suppose particle $\v x^{(j)}$ is into consideration. Particles that are far away, $\parallel \! \v x^{(l)} -\v x^{(j)} \! \parallel^2 \gg h$, do not contribute to the two terms in \reff{gradKL1}. On the other hand, if there is an $l$-th particle that is within the kernel scale of $\v x^{(j)}$, the $j$-th particle will be driven by a force from the second term in \reff{gradKL1} given by, 
\mi
\v x^{(j)}_{k,i}= \v x^{(j)}_{k,i-1} + \gamma (\v x^{(j)}_{k,i-1}-\v x^{(l)}_{k,i-1}),
\mf 
where $\gamma=\frac{K(\v x^{(l)}_{k,i-1},\v x^{(j)}_{k,i-1})}{h}>0$  so that the force given by this term is repulsive. It tends to separate the particles that are within the kernel scale.

\subsection{Importance sampling and the variational mapping particle filter}\label{sec:impsam}

\ma{The variational mapping sampling scheme is suitable to be coupled with importance sampling. The last intermediate density obtained with the variational mapping sampling scheme can be used as a proposal density of the posterior density. Suppose we denote as $\v x_k^{(1:N)}\doteq \v x_{k,I}^{(1:N)}$ to the particles obtained in the last mapping iteration. The values of the proposal and posterior densities at the particle positions are used to estimate the weights}
\mi
\tilde w^{(j)}_k= \frac{p(\v y_k|\v x_k^{(j)}) p(\v x_k^{(j)}|\v y_{1:k-1})}{q(\v x_k^{(j)})}, \label{weights}
\mf
and then they are normalized with $w^{(j)}_k = \tilde w^{(j)}_k/\sum_j^{N_p}\tilde w^{(j)}_k.$

Since the output of the filter is a set of weighted samples, the sequential posterior density (e.g. \reff{anaPost})  has to include the weights of the previous cycle,
\begin{align}
  p(\v x_k^{(j)}|\v y_{1:k}) &\propto  p(\v y_k|\v x_k^{(j)}) \sum_l w^{(l)}_{k-1} p(\v x_k^{(j)}|\v x_{k-1}^{(l)}). \label{posteriorWeights}
\end{align}

To compute the weights between the posterior density and the proposal density --final density of the MPF--, we also need to evaluate the proposal density at the particle locations. We require the proposal density to be known (apart from the set of samples that we have from that density). There are two ways to obtain it. One way is to use kernel density estimation with the set of samples, $\v x_k^{(1:N)}$, using the same kernels as in the mapping.  The second option is to determine at each mapping iteration how the density is transformed with the mapping (e.g. \cite{moselhy12}).

\pj{Before the mapping, we choose \ma{as initial intermediate} density, for instance, an equally weighted Gaussian mixture centered at $\dynm(\v x_{k-1}^{(i)})$ and with covariances $\v Q$, }
\begin{align}
  q(\v x_{k,0}|\v y_{1:k}) &\propto \sum_{i=1}^{N_p} \exp \left[-\frac{1}{2}\norm{\v x_{k,0}-\dynm(\v x_{k-1}^{(i)})}_{\v Q}^2 \right].
\end{align}

In the following mapping iterations $i>0$, the density changes according to the mapping $T$,
\mi
q_T(\v x^{(j)}_{k,i}) = \frac{q(\v x^{(j)}_{k,i-1})}{\det{\nabla_x T}} = \frac{q(\v x^{(j)}_{k,i-1})}{|\v I - \epsilon \mathbb H|}, \label{qTransform}
\mf
where $|\v I - \epsilon \mathbb H|$ is the determinant Jacobian of the transformation and $\mathbb H$ is the Hessian of the Kullback-Leibler divergence. From \reff{gradKL1}, the Hessian is given by 
\mia
\mathbb H&=& \frac{1}{N_p}\sum_{l=1}^{N_p} \left\{\nabla_{\v x^{(j)}_{k,i-1}} K(\v x^{(l)}_{k,i-1},\v x^{(j)}_{k,i-1})  \left[\nabla_{\v x} \log p(\v x^{(l)}_{k,i-1})\right]^\transp\right.\nonumber\\  & +& \left.\nabla_{\v x^{(j)}_{k,i-1}} \nabla_{\v x^{(l)}_{k,i-1}}  K(\v x^{(l)}_{k,i-1},\v x^{(j)}_{k,i-1})\right\}.\label{nablaphi}
\mfa

The calculation of the Hessian of the Kullback-Leibler divergence, \reff{nablaphi} is computationally demanding. Although note that the second derivatives are well known for standard kernel functions and the symmetry of the kernel can be used to reduce the calculations. However, the Hessian calculation is required at each mapping iteration to evolve $q_{k,i}^{(j)}$  from the previous intermediate density $q_{k,i-1}^{(j)}$.

\ma{The weights account for the bias introduced in the filter when the set of samples obtained by the sequence of mappings is not exactly distributed according to the target density. }\pj{However, it should be kept in mind that our estimate of the posterior density \reff{posteriorWeights} is not exact, so it is unclear what the calculated weights actually mean. It would be interesting to \ma{determine the accuracy} of this estimate, but that is beyond the scope of this work.}%

The weights are also  useful as a {\em diagnostic} of the quality of the convergence of the optimization method. In that case, an evaluation of the weights for the current intermediate density is conducted in each mapping iteration. If the given weights give an effective sample size which is over the threshold no further mapping iterations are required. In the conducted experiments of this work, in general, the effective sample size was over 98\%  after 50 mapping iterations during the whole recursion. In other words, if the number of mapping iterations is large enough $\ge 50$, the weights will be almost equal. In high-dimensional systems, a convergence criterion based on the module of the gradient of Kulback-Leibler divergence can be used. This avoids the calculations of the Hessian or kernel density estimations. The convergence of the technique is evaluated through experiments with both measures in Section \ref{subsec:optim}. 

\subsection{Experiment details}
The Lorenz-63 system is given by
\begin{align}
&\frac{\ud x}{\ud t}=\sigma (y-x), \nonumber\\
&\frac{\ud y}{\ud t}=x  (\rho-z)-y, \\
&\frac{\ud z}{\ud t}=x y - \beta z. \nonumber 
\end{align}

The equations are solved using a fourth-order Runge-Kutta scheme. The parameters are chosen at their standard values $\sigma=10$, $\rho=28$ and $\beta=8/3$. The integration time step is $0.001$ and the assimilation cycles are every $0.01$ (10 integration time steps). The observation operator  $\mathcal{H}$  is taken as the identity matrix, for the full-observed state experiments in Section \ref{sec:results}. Further experiments are conducted in Section \ref{subsec:partial} using partial observations. The model error covariance $\v Q$ is chosen diagonal. Its diagonal elements are set to 30\% of the climatological variance of the variable, e.g. $Q_{11}= 0.3 \sigma_x^2$, where $\sigma_x^2$ is the time-series variance of variable $x$. The observation error covariance is $\v R=0.5 \v I$. \linelabel{ln-kde} \ma{The Lorenz-63 experiments use importance sampling to evaluate the bias introduced by the sampling.} \pj{The weights} are evaluated using kernel density estimation \pj{with} radial basis functions in the state space in coherence to the ones used for the mapping (same Gram matrix). 

For the 40-variable Lorenz-96 system experiments, the set of equations is
\mi
\frac{\ud x_i}{\ud t}=(x_{i+1}-x_{i-2}) x_{i-1} - x_i + F, 
\mf
where $i=1,\cdots,40$ and cyclic boundary conditions are imposed, $x_0=x_{40}$, $x_{-1}=x_{39}$ and $x_{41}=x_1$. The equations were integrated using a fourth-order Runge–Kutta scheme, with an integration step of 0.001. The forcing is $F=8$ which results in chaotic dynamics. The observational time resolution is $0.05$.\linelabel{rev-L96q} \ma{The observational error covariance is $\v R=0.5 \v I$. The model error covariance is $\v Q=0.3 \v I$. This amplitude of the model error noise is representative for the two-scale Lorenz 96 system which was estimated to be 0.3 using information measures by \cite{pulido17} and 0.3-05 using an Expectation-Maximization algorithm coupled to an ETKF\cite{pulido18}.}  The ensemble is composed of 20 particles. The initial ensemble particles are states taken randomly from a climatology. \ma{Because of the large dimension, we do not use  the effective number of particles via kernel density estimation as in the Lorenz-63 experiments to evaluate convergence of the optimization\pj{, but the module} of the gradient of the Kullback-Leibler divergence.}

The compartment stochastic dynamical model for cholera is the one throughly described in \citep{ionides06}. A short description is given here. The individuals are classified as susceptible ($S$), infected individuals ($I$), while recovered individuals belong to three classes ($R^{1:3}$) to allow for different inmune periods. The equations for the number of individuals for each category are given by
\begin{eqnarray}
  dS_t&=&dN_t^{BS}-dN_t^{NI}-dN_t^{SD}+dN_t^{R^kS} \nonumber \\
  dI_t&=&dN_t^{SI}-dN_t^{IR^1}-dN_t^{IC}+dN_t^{ID}\nonumber \\
  dR^1_t&=&dN_t^{IR^1}-dN_t^{R^1R^2}-dN_t^{R^1D}\label{si3r} \\
  dR^2_t&=&dN_t^{R^1R^2}-dN_t^{R^2S}-dN_t^{R^2D}\nonumber \\
  dR^3_t&=&dN_t^{R^2R^3}-dN_t^{R^3S}-dN_t^{R^3D} \nonumber
\end{eqnarray}

Transmission is given by a stochastic differential equation
\mi
dN^{SI}_t = \lambda_t S_t dt + \epsilon I_t S_t/P_t dW_t, \label{transmission}
\mf
where $dW_t$ is a Gaussian white noise. The transitions between categories are given by
\begin{eqnarray}
  &dN_t^{IR^1}=\gamma I_t dt, \qquad &dN_t^{R^{l-1}R^l} = r k R^{l-1}_t dt,\nonumber\\
  &dN_t^{R^lS}=rk R^k_t dt, \qquad &dN_t^{SD} = m S_t dt,\nonumber\\
  &dN_t^{ID} = m I_t dt, \qquad &dN_t^{R^lD}= m R^l_t dt,\\
  &dN_t^{IC}=m_cI_t dt, \qquad &dN_t^{BS}= dP_t + m P_t dt,\nonumber
\end{eqnarray}
where $l=2,3$, $B$ represents birth, $C$ is cholera mortality and $D$ denotes death from other causes. Equations {\bf \ref{si3r}} are integrated using an Euler-Maruyama scheme with time step of $1/20$~month. The observations are cholera mortality data which are given by
\mi
y_t \sim \mathcal{N}\left[N_t^{IC} - N_{t-1}^{IC}, \tau^2 (N_t^{IC} - N_{t-1}^{IC})^2\right].
\mf

\pj{ While the mapping particle filter can handle non-additive model errors we decided to transform the system of equations above to allow for an additive model error implementation.} We use an augmented state space defining a new state variable $T$ as $dT_t=dW_t$, so that the augmented system has additive model error whose the gradient of the log-posterior density is represented by \reff{anaPost}. To add diversity  to the particles in the filter a small additive noise term is added in all the equations. The variance of this additive noise is set to 10\% of the $dW_t$ variance. This noise is not used to generate the true trajectory, it is only included in the particle filter.

\subsection{The stochastic optimization methods and their convergence}\label{subsec:optim}

An optimization of the Kullback-Leibler divergence is conducted in the mapping particle filter \pj{to push the prior particles to samples from the posterior density.} The particles are driven  in a purely deterministic transformation by the variational mapping from the prior density to the posterior. The intermediate densities \ma{are represented through} the particles. Each particle is then moved along the direction of the steepest descent of the Kullback-Leibler divergence, which considers all the particle positions. 

Several stochastic gradient-based optimization methods have been recently developed in the machine learning community \citep{zeiler12,kingma15} which are directly applicable to the mapping particle filter. They are focused on efficient convergence in high-dimensional control spaces when the cost function is noisy. The source of this noise is subsampling. These optimization methods, and their succesful convergence rates, have been instrumental for the success of deep learning applications (e.g. \citealt{lecun15}). The variational mapping in this work is based on stochastic gradient-based optimization methods.

The stochastic gradient descent method has a fixed learning rate for all the iterations and all the control variables. This is well-known to cause some inconveniences, in particular along extended valleys where zig-zag convergence arises. As the gradient decreases in some directions, the convergence along those variables is very slow. The adaptative schemes search for a dynamic learning rate for each control variable. Ideally, a Newton method would be the optimal learning rate if the optimization problem is quadratic, but it requires Hessian evaluations which are  computationally prohibitive in high-dimensional spaces. \linelabel{rev-adadelta} The learning rate for each control variable \ma{in the stochastic optimization method, adadelta \citep{zeiler12}}, is based in a global learning rate over a running average of the past absolute directional derivatives along the direction of the control variable. Therefore, the learning rate increases for directions with small directional derivatives. The method improves substantially but still some stagnation after the first iterations has been reported. To overcome this weakness, the algorithm \ma{for stochastic optimization} adam\citep{kingma15} estimates the first and second moments of the gradients with bias correction. These two \ma{stochastic optimization} algorithms, adadelta and adam, together with a descent gradient method were evaluated in the experiments. \linelabel{ln-learningRate}\ma{The chosen initial learning rate of the optimization methods is a tradeoff between convergence speed and smoothness flow assumptions.  The recommended values exhibited a good performance, so that no manual tuning  of the learning rate was required\citep{zeiler12}.  }

\begin{figure}
\hspace*{-.3cm}\centering\includegraphics[width=3.4in]{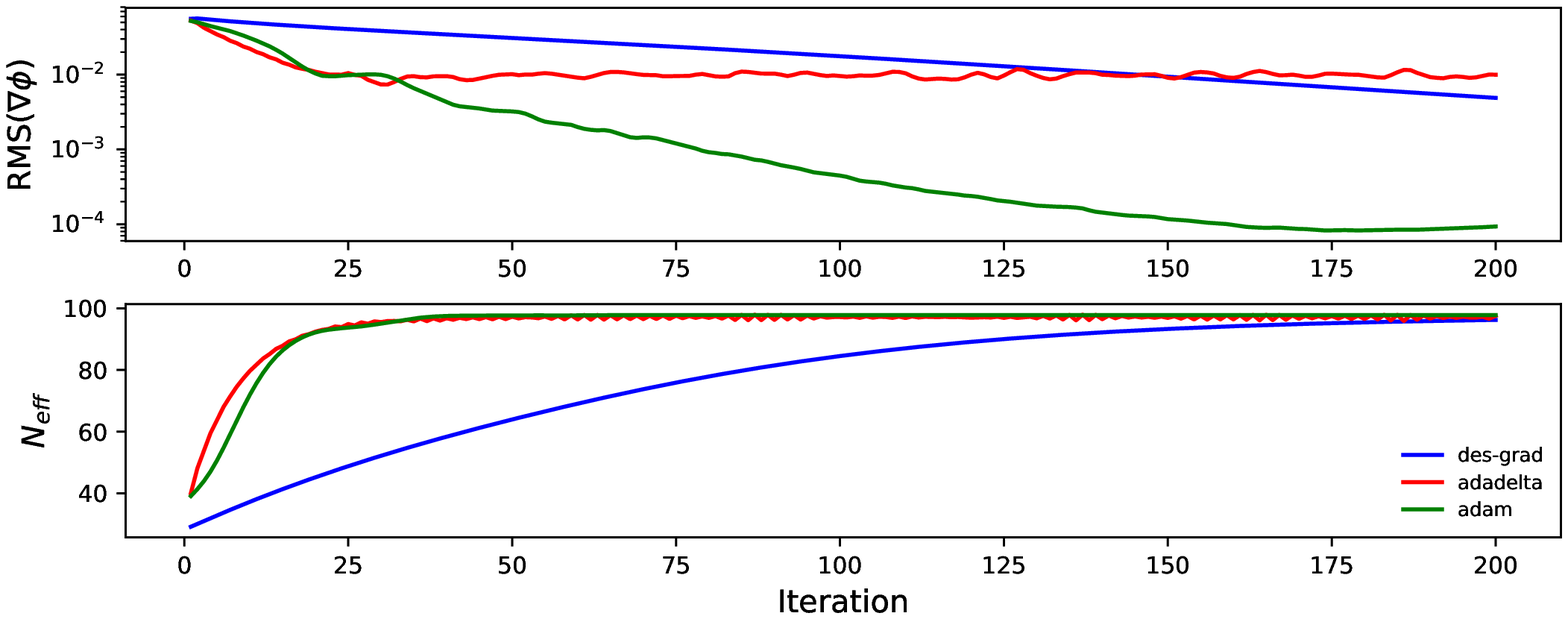}
\caption{The root-mean-square of the $KL$  gradient as a function of the mapping iteration for the descent gradient (blue line), adadelta (red line) and adam (green line) optimization algorithms (upper panel). This corresponds to the first assimilation cycle in which the three optimization algorithms share the same prior density. The number of effective particles as a function of the mapping iteration for $N_p=100$ (lower panel). }\label{conv1}
\end{figure}

\begin{figure}
\hskip -.4cm
\centering  \includegraphics[width=3.45in]{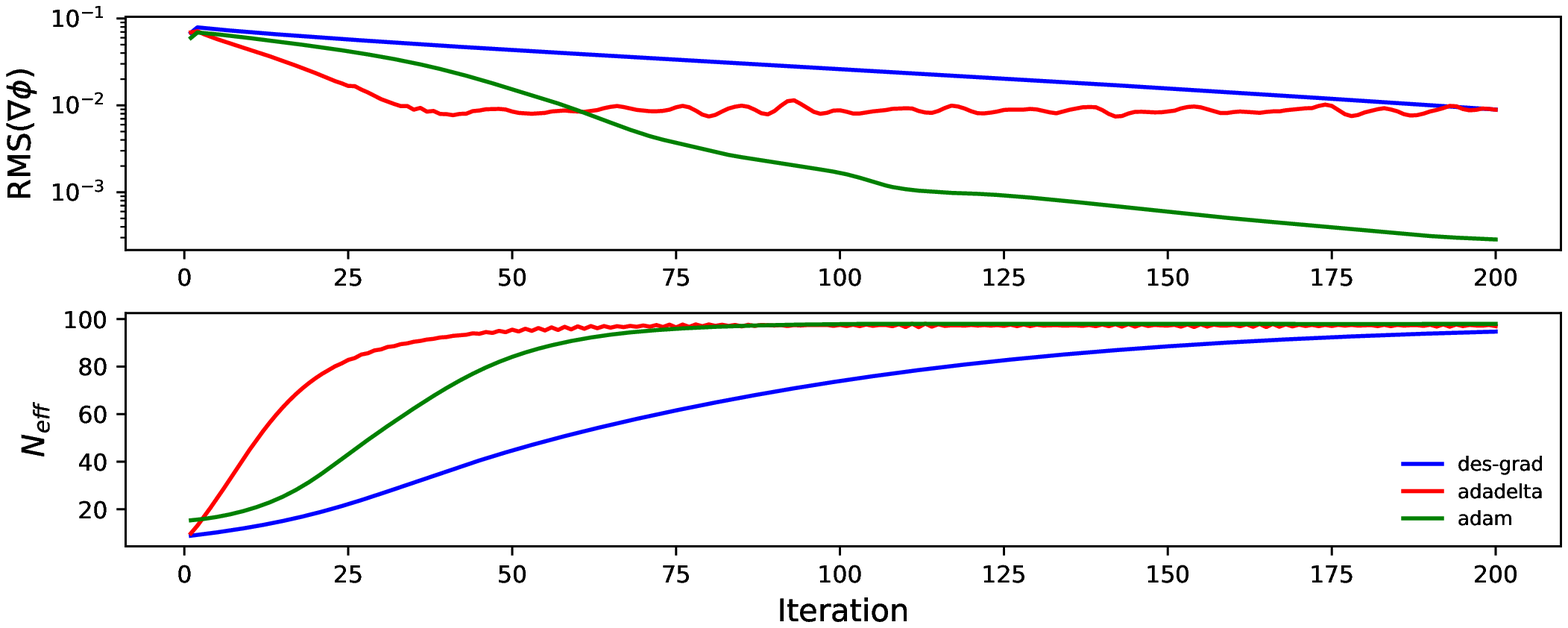}
\caption{As Fig. \ref{conv1} but for the 10-th time iteration, so that the prior density for the optimization algorithms is not the same.}\label{conv2}
\end{figure}

The convergence of the variational mapping is shown in Fig. \ref{conv1} as a function of the mapping iteration at the first assimilation cycle. At the first cycle, the three optimization methods, descent gradient, adadelta and adam, have the same initial density. Upper panels of Fig. \ref{conv1} show the root-mean-square of the KL gradient, and lower panels the number of effective particles. The adadelta converges fastest, close with adam. For a larger number of iterations, the gradient of KL divergence is still diminishing for adam but almost no change after 25 iterations for adadelta. However, the number of effective samples does not differ between them, 40 iterations are required to achieve the maximum--98\% of the total number of particles. The descent gradient requires about 200 iterations to achieve this number of effective particles.

Figure \ref{conv2} shows the convergence at the 10th assimilation cycle of the optimization methods for evaluating the recursive effects. Notably, adadelta has the fastest convergence with the minimum root-mean-square gradient similar to the one obtained in the first iteration. On the other hand, adam converges slowly in the first mapping iterations and starts to increase the convergence rate after 20 iterations (an effect that is likely caused by the momentum equations in adam which require some iterations to ``warm up''). Adadelta has reached the maximum number of effective particles in 50 iterations. Because of the fast convergence and because the effective particle numbers is the most relevant parameter for the particle filter we took  adadelta with 50 iterations and \linelabel{ln-003}\ma{an initial learning rate of 0.03} as the default setting for the experiments.

\linelabel{ln-KL-weights}\ma{There is a clear correlation between the decrease of the root-mean-square of the KL gradient and the increase in the number of effective particles in Fig. \ref{conv2} for the three stochastic optimization methods. This implies as mentioned that both measures could be used to determine the convergence in the optimization and the required number of iterations for a given threshold. Along this line, we note that the Kullback-Leibler divergence can be directly expressed in terms of the weights.} %

\ma{Using Monte Carlo integration of the Kullback-Leibler divergence,}
\mi
KL(q\lVert p)= \frac{1}{N_p}\sum_{j=1}^{N_p} \log\left[\frac{q(\v x_k^{(j)})}{p(\v x_k^{(j)}|\v y_{1:k})}\right],
\mf
from \reff{weights}, it results
    \mi
KL(q\lVert p)= - \frac{1}{N_p}\sum_{j=1}^{N_p} \log\left(w^{(j)}_k N_p\right). \label{KLweights}
\mf
As expected if the weigths are equally distributed in the particles, the Kullback-Leibler divergence is zero, then $q(\v x_k)\approx p(\v x_k|\v y_{1:k})$. In other words, as the number of effective particles $N_{eff}$ tends to $N_p$, the Kullback-Leibler divergence tends to zero, its minimum.

\subsection{Experiments on sensibility of the MPF}\label{sec:bias}

\begin{figure}
\hskip -.4cm
\centering\includegraphics[width=4.20in]{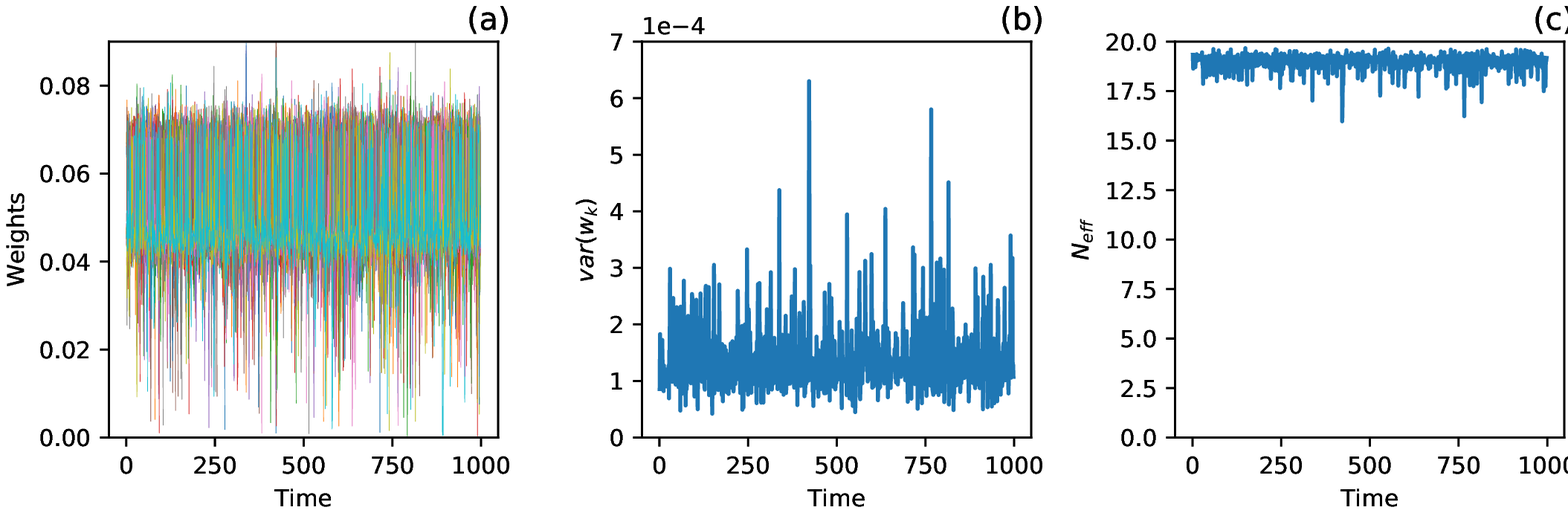}

\caption{(a) Evolution of the weights of all the particles for the Lorenz-63 system in a long time sequence for $I=50$ and without threshold.  Because the weights of all the particles are plotted, the last particle plotted is the most visible one.  (b) The variance of the weights as a function of time. (c) Number of effective particles.}\label{weightevol}  
\end{figure}

\begin{figure}
\hskip -.4cm
\centering\includegraphics[width=4.20in]{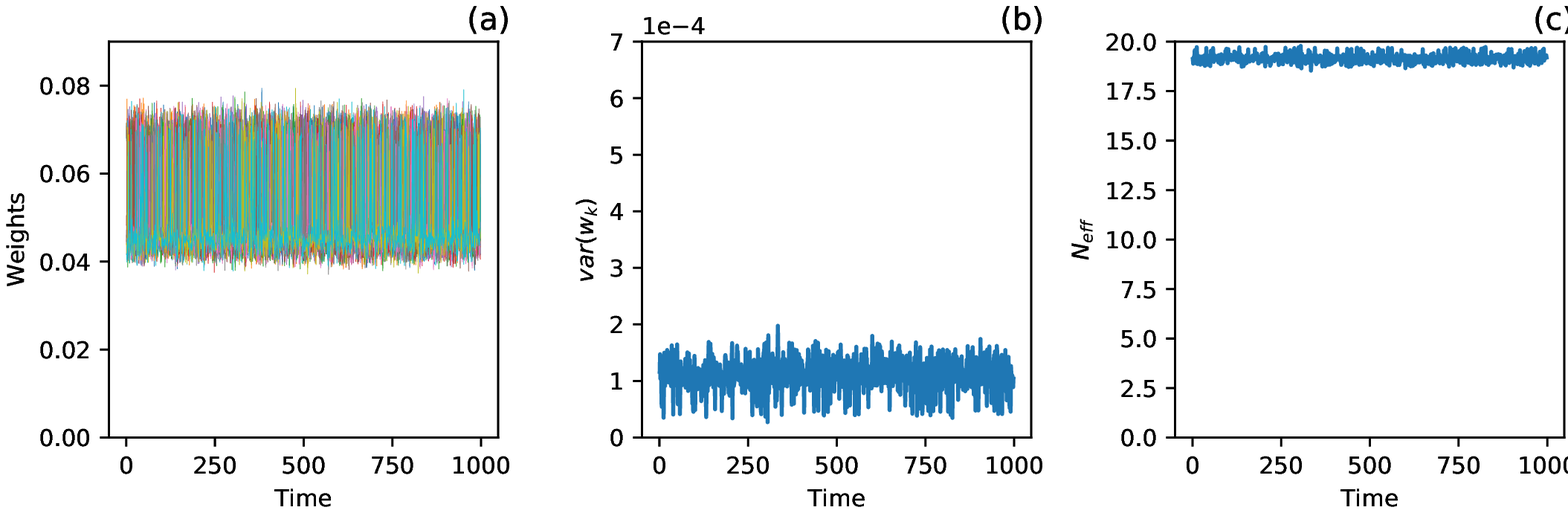}
\caption{Idem Fig. \ref{weightevol} for $I=100$ and without threshold.}\label{weightevol2}  
\end{figure}

\ma{The performance of the mapping particle filter can be evaluated through the weights of the particles.  As a function of the mapping iterations of the filter, the sequence of mappings is expected to start with a set of particles with most of the particles with almost zero weights and a few particles with very large weights. Note from \reff{KLweights} that particles with almost zero weights produce a large Kullback-Leibler divergence. Then, as the particles are pushed toward the posterior density, their weights will tend to be equally distributed \pj{for} most of the particles. If the variational mappings work effectively, only a few particles should remain with low weights. In other words, the variance of the weights should be small for all the cycles. Figure \ref{weightevol}a shows the evolution of the weights of all the particles as a function \pj{of time} for $I=50$ and without threshold for the Lorenz-63 system. Most of the particle weights are concentrated around 0.05, between 0.04 and 0.07 ($N_p=20$ particles). There are some cycles in which one particle \pj{obtains a} low weight but in the following cycle the weight of that particle is again within the equally-weighted range. \pj{Note that we did not perform any resampling.} The variance of the weights is around $1.25~10^{-4}$ (Figure \ref{weightevol}b). The number of effective particles is in general about 19 (from a total of 20 particles), but in a few cycles it can descend down to 16.  A further experiment was conducted in which the maximum number of iterations was increased to 100. In that case, all the particles remain with weights larger than 0.04 during the whole time sequence and the peaks of variance of the weights found in Fig. \ref{weightevol}b are not present (see Fig. \ref{weightevol2}). The number of effective particles is in this case over 18 for all the cycles}

\begin{figure}
\hskip -.4cm
\centering\includegraphics[width=4.20in]{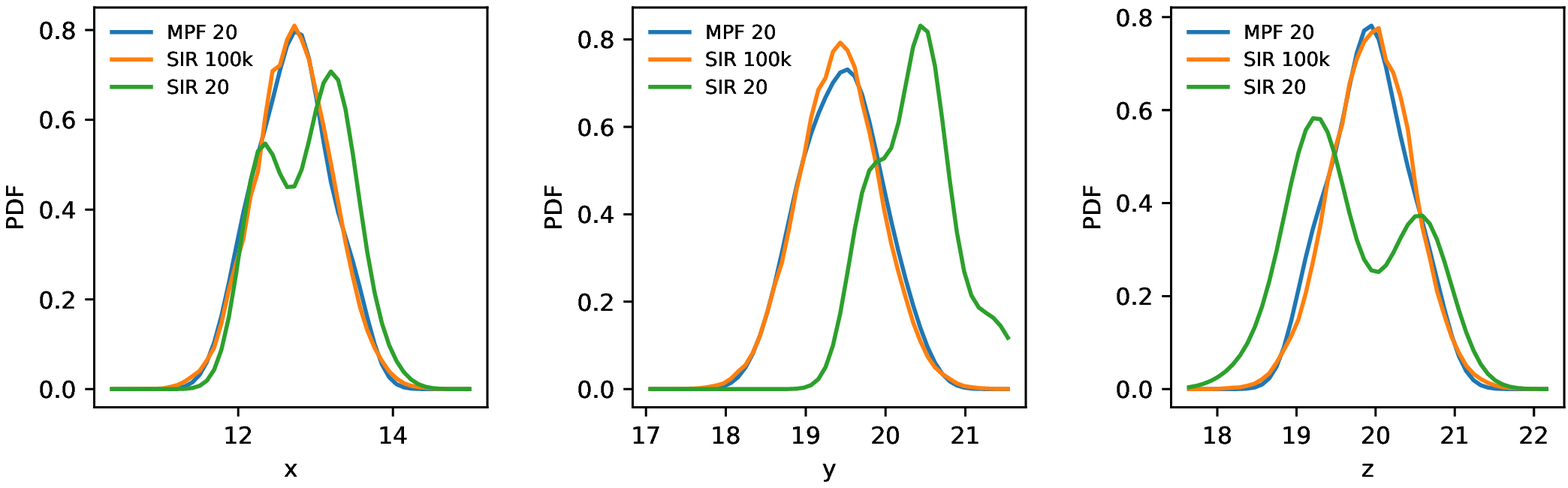}

\hskip -.4cm
\includegraphics[width=4.20in]{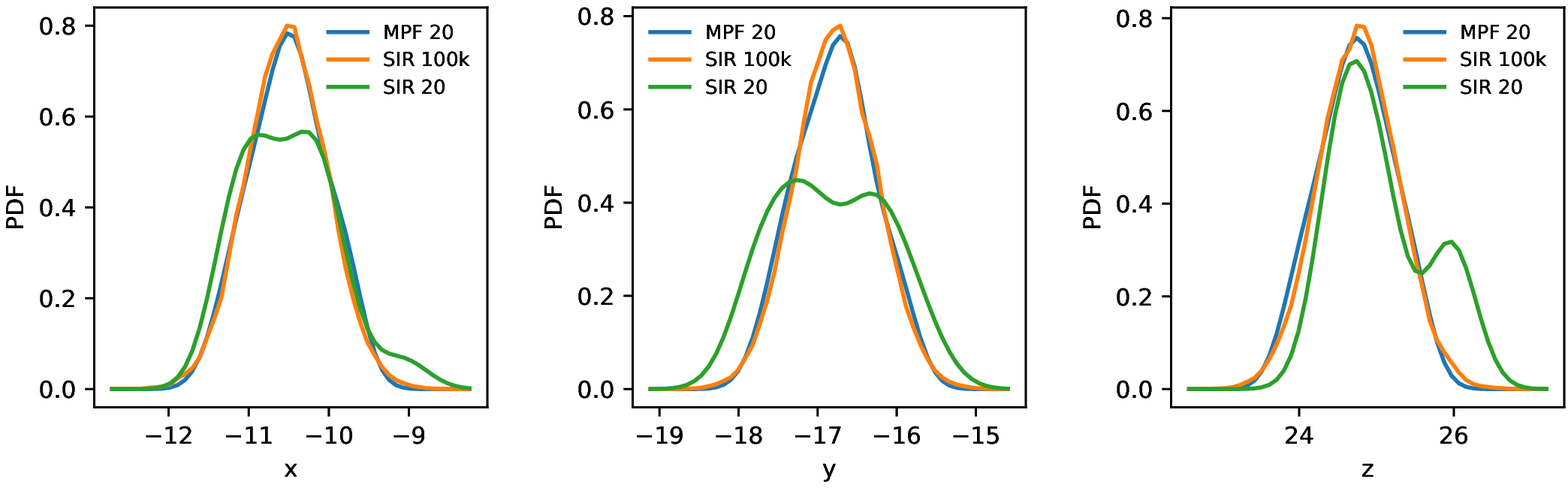}
\caption{Marginalized posterior densities in each variable of the Lorenz-63 for a SIR with 100,000 particles (red line), SIR with 20 particles (green line) and MPF with 20 equally weighted particles (blue line) at the 450th cycle (upper panels) and 500th cycle (lower panels).}\label{truepos}  
\end{figure}

\ma{To evaluate the MPF without using importance sampling, we conducted a Lorenz-63 experiment without using weights in the particles after the mapping. As criterion of convergence we use  $|\nabla KL_i|/|\nabla KL_0|<0.07$ averaged over all the particles. The MPF experiment uses 20 particles and is compared with a SIR filter using 20 particles and another with 100,000 particles. The last is expected to be close to the true posterior density. Figure \ref{truepos} shows the marginalized posterior densities of the three experiments in each Lorenz-63 variable. Upper panels correspond to cycle 450th and lower panels to cycle 500th. The marginalized densities given by the MPF with 20 particles are remarkably close to the SIR experiment with 100,000 particles after a long sequence of cycles, without using resampling nor importance sampling. On the other hand, the SIR filter with 20 particles has large differences with the true posterior density not only in the mode but also the uncertainties are larger than the true marginalized posterior density in some directions and smaller in other ones. Because of the resampling step some tendency of the SIR with a small number of particles to generate a bimodal marginalized densities in some variables was also found.}

\begin{figure}
\hskip -.4cm
\centering\includegraphics[width=3.650in]{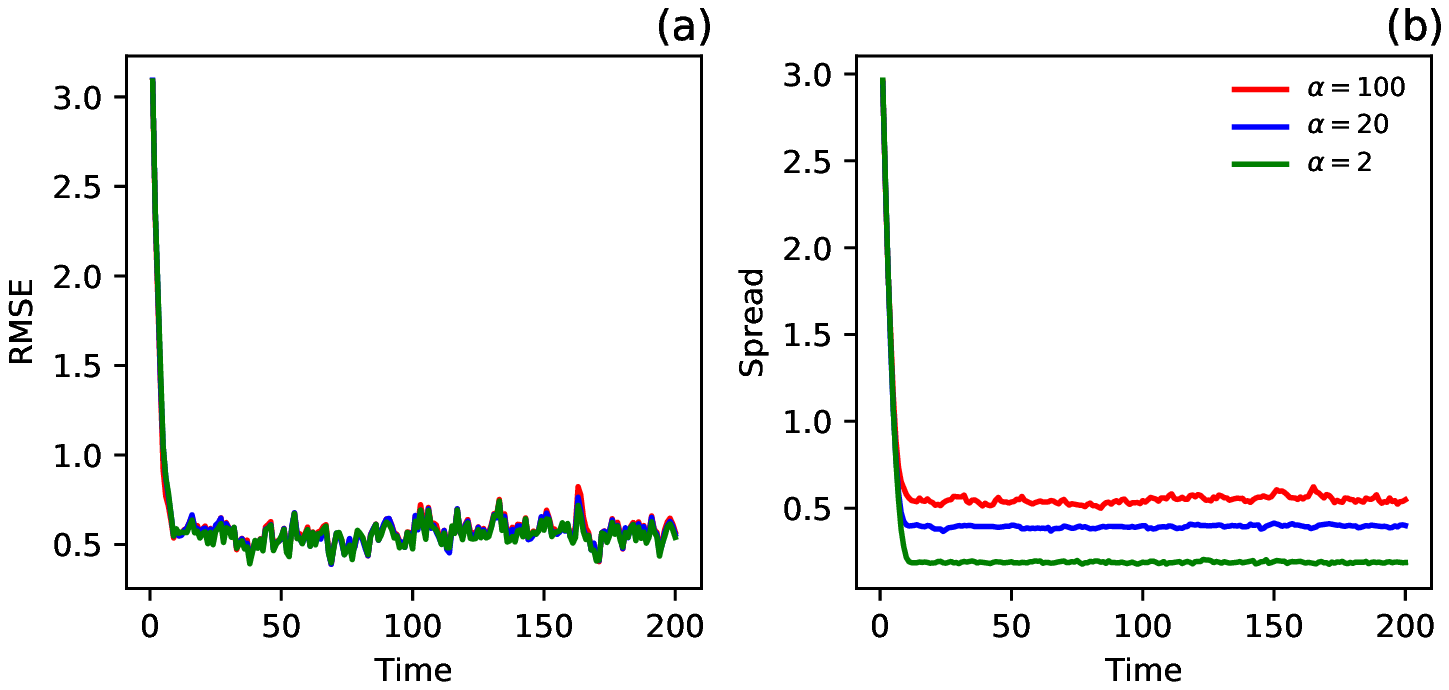}
\caption{(a) RMSE for the MPF in the Lorenz-96 experiments using a bandwidth of $\alpha=2, 20, 100$. (b) The corresponding spread of the particles of the posterior density. }\label{l96bandw}
\end{figure}

\linelabel{ln:alpha}
The radial basis functions used as kernel require to set the bandwidth, which in this work was chosen as $\v A=\alpha \v Q$. The parameter $\alpha$ depends on the number of particles and the dimensions of the problem \pj{and} requires some tuning. The optimal parameter will be the one that produces the \ma{sample} spread \ma{equal to the spread of} the sequential posterior density. The bandwidth parameter $\alpha$ used in the Lorenz-63 experiment with 20 particles is $\alpha=1$. \pj{For} the experiment with 100 particles we use $\alpha=0.5$. Figure \ref{l96bandw} shows experiments for the 40 variables Lorenz-96 system in which we vary the bandwidth from $\alpha=2$ to $\alpha=100$. The RMSE shows only slight changes with similar values in the three experiments (Figure \ref{l96bandw}a). The RMSE measure for the MPF appears rather robust to the kernel bandwidth. Figure   \ref{l96bandw}b shows the spread of the experiments for different $\alpha$ values. A choice of $\alpha=20$ appears to give an optimal spread of the particles for this experiment.

\subsection{Experiments using partial observations}\label{subsec:partial}

\begin{figure}
\hskip -.4cm
\centering\includegraphics[width=3.650in]{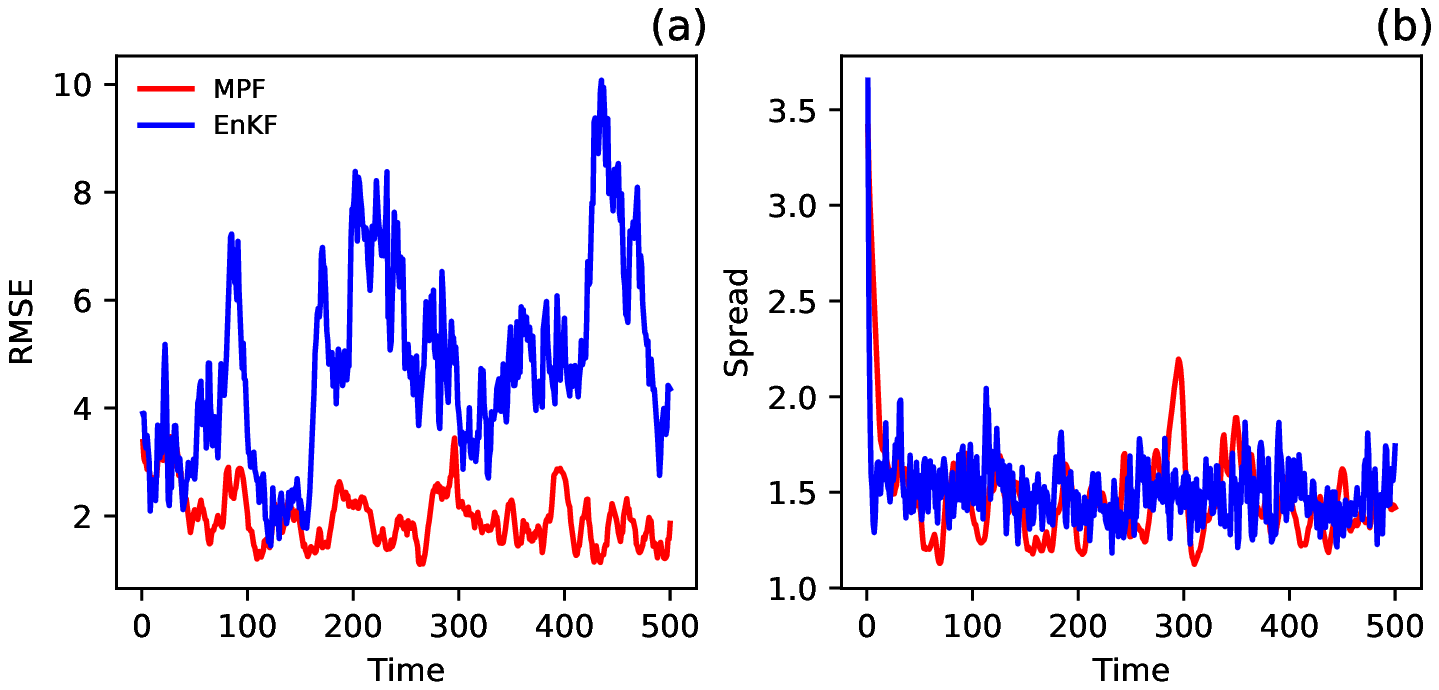}
\caption{RMSE (a) and spread (b) of the state as a function of the assimilation cycles produced by  EnKF and MPF  in the 40-variables Lorenz-96 system with 20 observations. Both filters use 20 particles.}\label{l96_20obs}
\end{figure}

\ma{An experiment to evaluate the performance of MPF for a partially observed Lorenz-96 system (40 variables and 20 observations) was conducted using 20 particles. Figure \ref{l96_20obs}a shows the RMSE given by the MPF and  the \pj{stochastic} EnKF without localization. The MPF outperforms the EnKF in terms of the RMSE measure. Figure \ref{l96_20obs}b shows the spread of the particles for both filters.   }

\begin{figure}
\hspace*{-.3cm}\centering\includegraphics[width=4.20in]{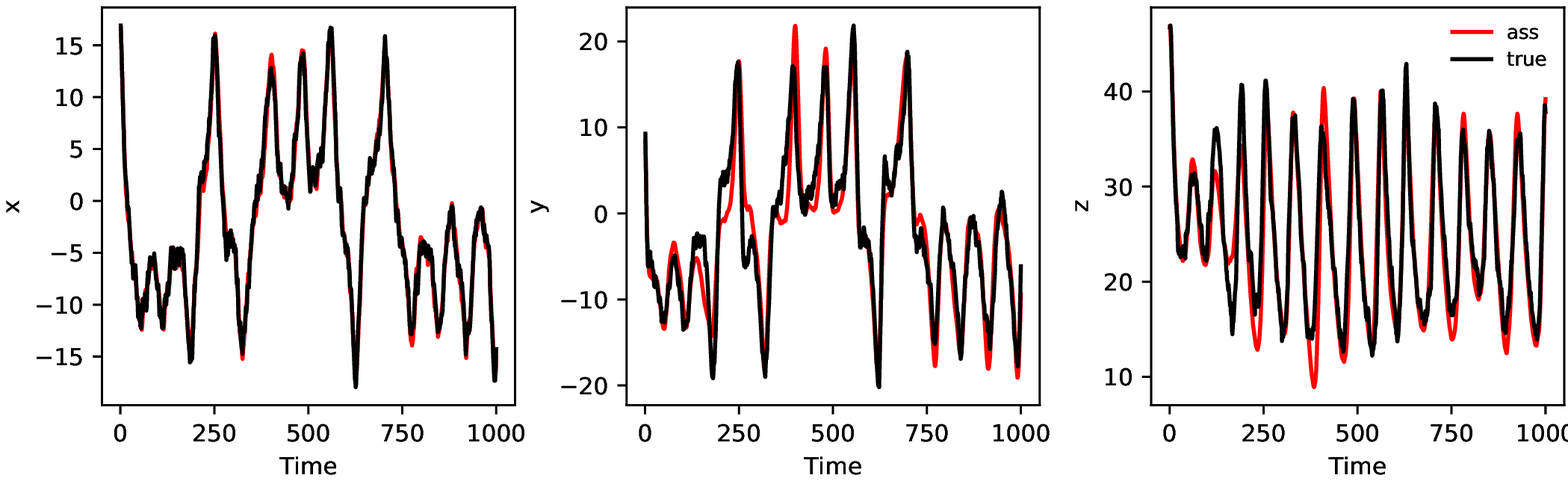}

\hspace*{-.3cm}\centering\includegraphics[width=4.20in]{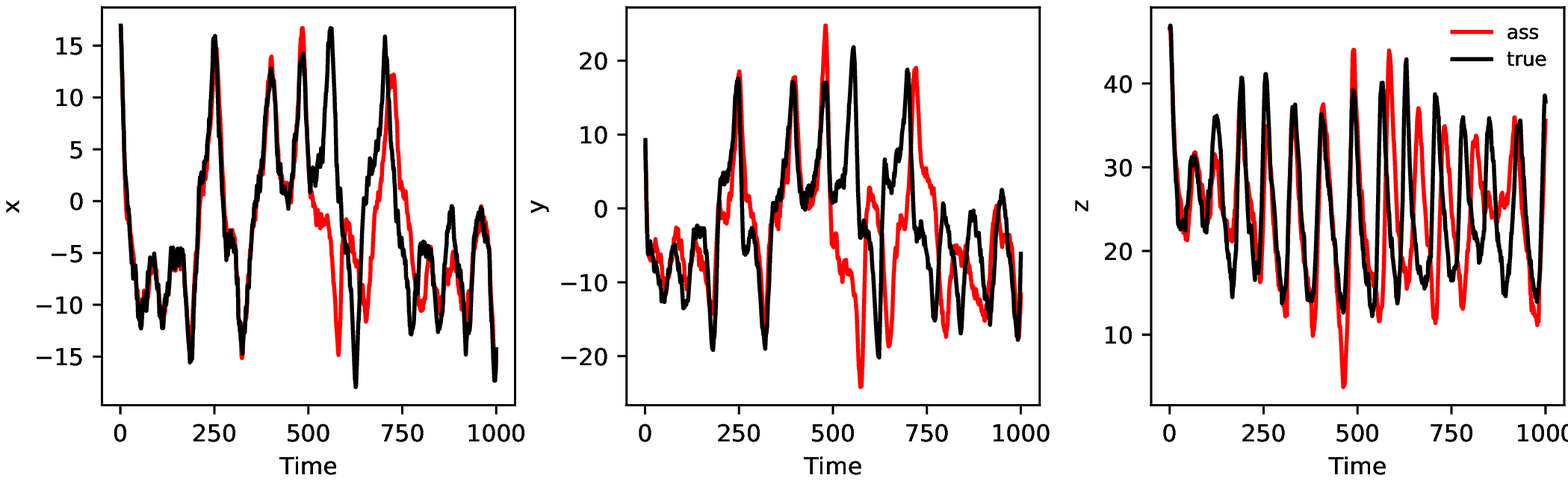}
\caption{Estimated mean state and the true evolution of the Lorenz-63 variables. The only observed variable is $x$, we  use only 5 particles. Upper panels show the MPF experiment and lower panels SIR experiment. Time axis corresponds to number of assimilation cycles, the length of the assimilation cycle is $\Delta t=0.01$. }\label{mpf_xobs-5npa}
\end{figure}

Figure \ref{mpf_xobs-5npa} shows the mean state in the three Lorenz-63 variables for the partial observation experiment. They are estimated by the particle filters using just 5 particles for the case in which only $x$ is observed and $\sigma_R=1.0$, $\v Q=0.3 \v I$. The MPF filter (upper panels) manifests a very good performance with the mean state estimate in the observed variable very close to the true value, while it is also capturing rather closely the behavior of the unobserved variables (Panels (b) and (c)). The information that is available in the observed variable is enough to recover through the prediction density correlations of the full hidden state. The SIR filter for the same conditions and same set of observations diverges (lower panels of Fig. \ref{mpf_xobs-5npa}), it finishes with all the weight  in only one particle.

\begin{figure}
\hspace*{-.3cm}\centering\includegraphics[width=4.20in]{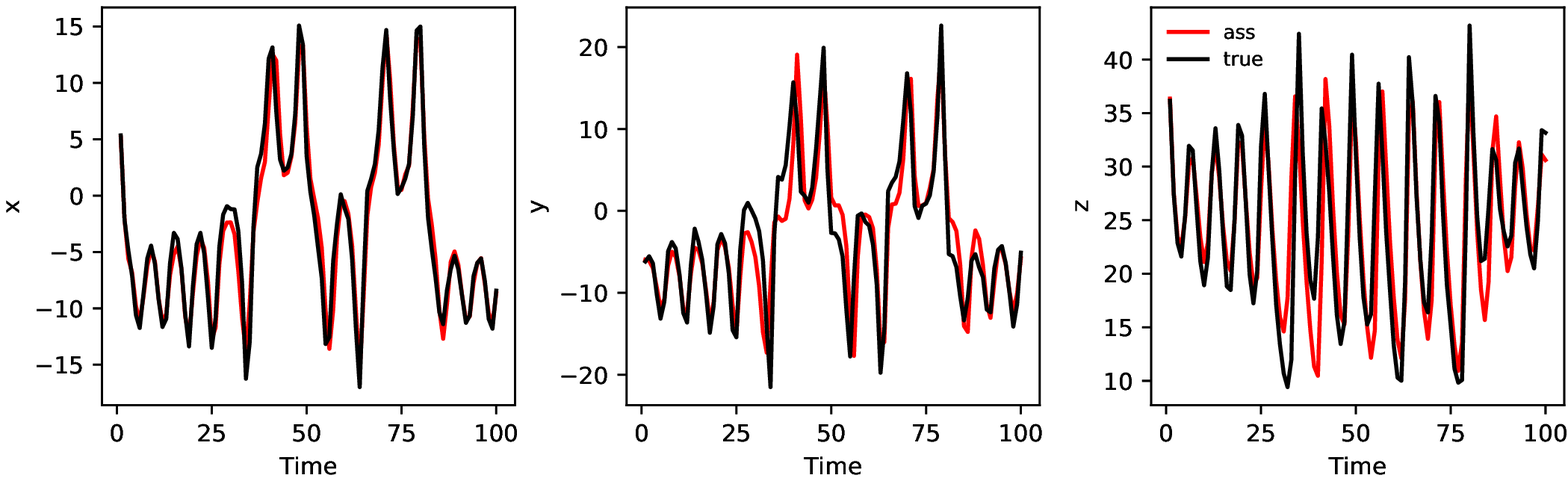}

\hspace*{-.3cm}\centering\includegraphics[width=4.20in]{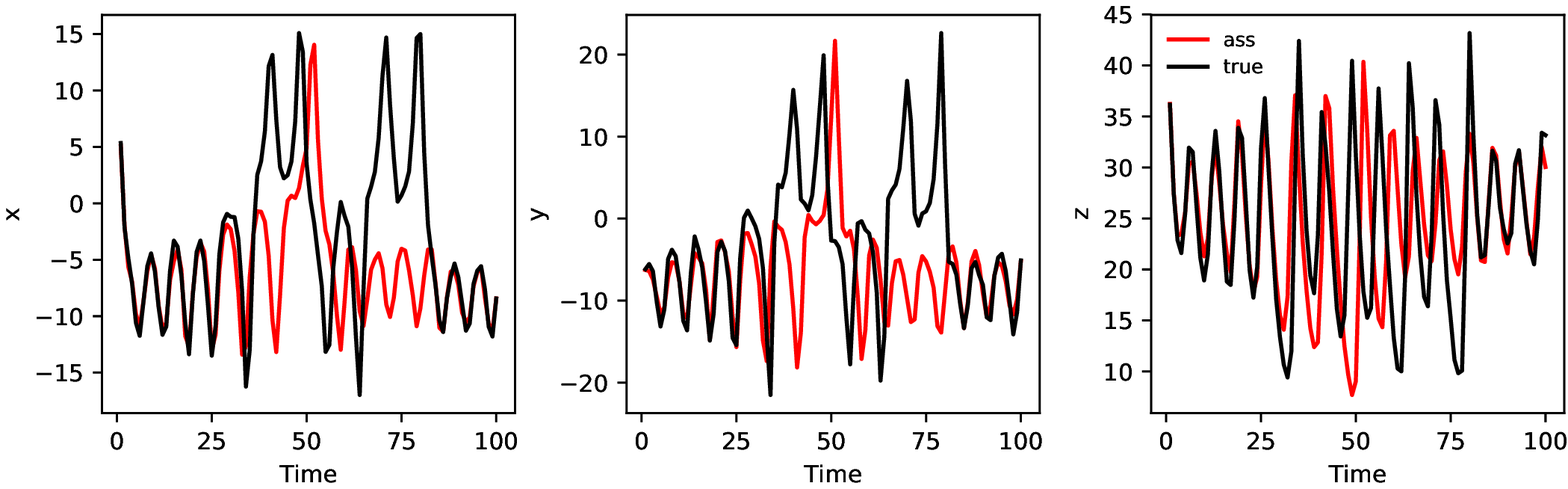}
\caption{Estimated mean state and the true evolution of the Lorenz-63 variables for a long assimilation window $\Delta t=0.1$. There is only one observed variable ($x$). A set of 20 particles is used in the filters.}\label{ass_mpf_xobs_20npa_dt0-1}
\end{figure}

As a further experiment, a more challenging one, we evaluate the performance of the MPF for long assimilation windows and partial observations, the assimilation cycle is extended to $0.1$. Note the integration step is kept as in all the experiments at $\ud t=0.001$ so that there are 100 model time steps between observations. The number of particles is 20. In this case, we expect a much stronger effect of the nonlinear terms in the Lorenz-63 equations. Figure \ref{ass_mpf_xobs_20npa_dt0-1} shows the mean state estimates with the MPF (upper panels) and SIR (lower panels). While the SIR filter diverges, the MPF exhibits a very good performance considering the small number of particles and partial sporadic observations. 

\begin{figure}
\hspace*{-.3cm}\centering\includegraphics[width=4.20in]{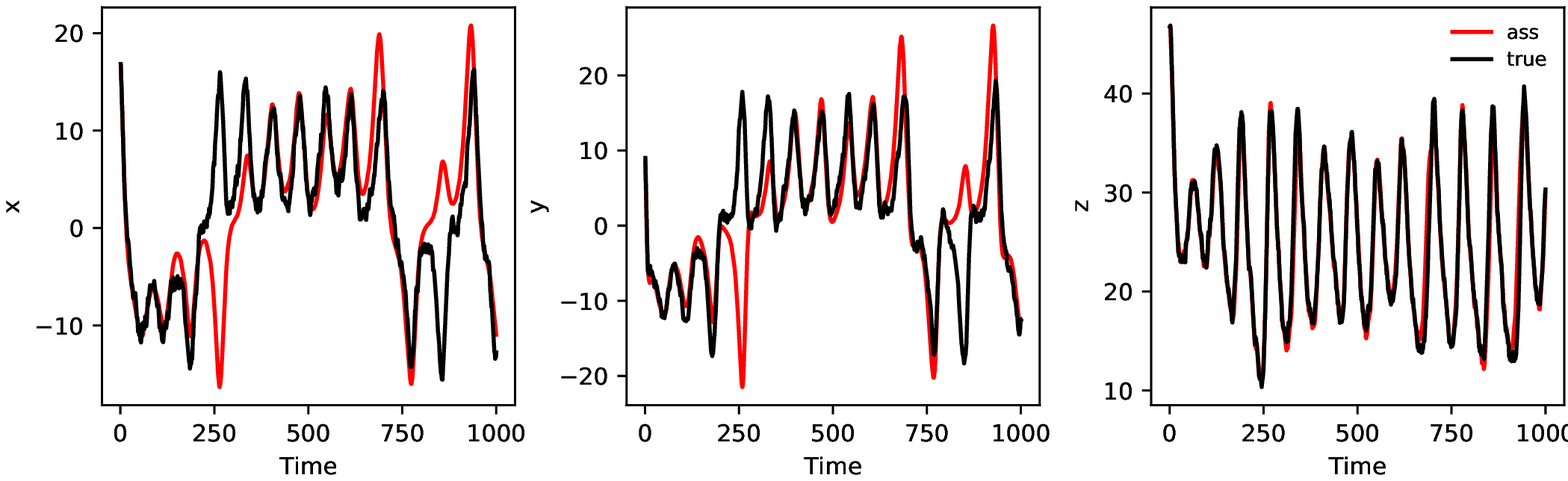}

\hspace*{-.3cm}\centering\includegraphics[width=4.20in]{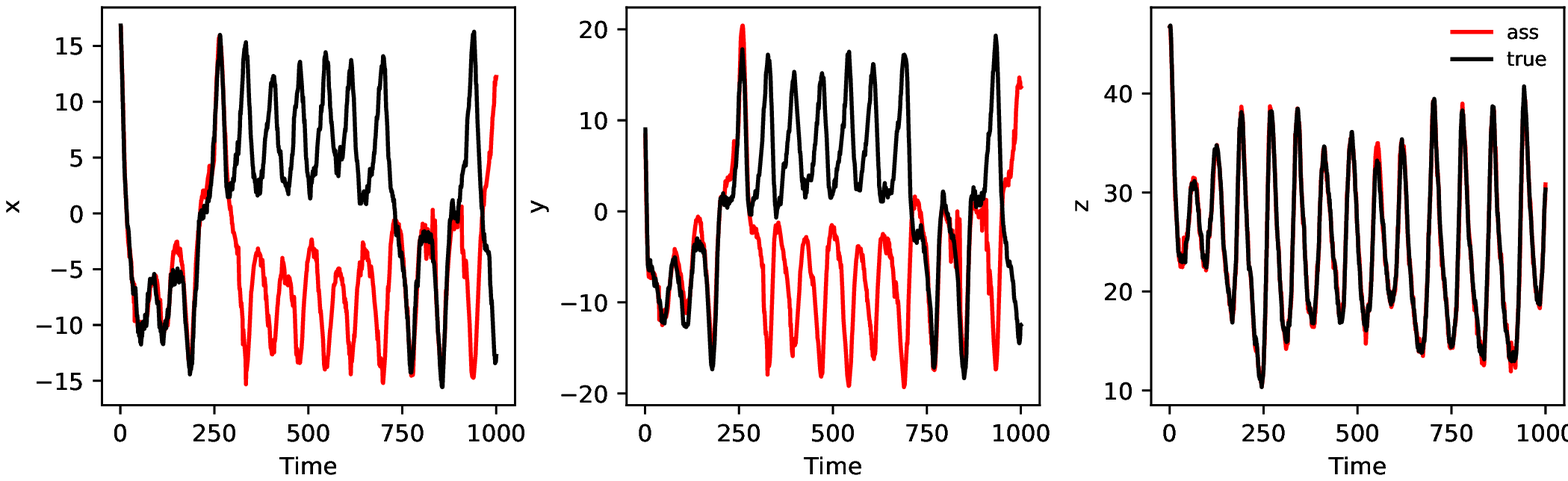}
\caption{Estimated mean state and the true evolution of the Lorenz-63 variables. The only observed variable is $z$. A set of 20 particles is used in the MPF. Upper panels show the estimated mean state with the MPF and lower panels with the SIR filter.}\label{ass_mpf_zobs-20npa}
\end{figure}

A third partial observation experiment is evaluated in which only $z$ is observed. The experiment with only $z$ as observed variable does not contain all the dynamical information from observations to recover the full state, since the variable $z$ does not contain information about the wing of the attractor that the other two variables $x, y$ are in.  We conducted an experiment with $\sigma_R^2=0.5$, $\v Q=0.15 \v I$ and 20 particles. Upper panels of Fig. \ref{ass_mpf_zobs-20npa} show the results for the MPF experiment. As expected, the filter is unable to recover the state in $x$ and $y$ variables for some assimilation cycles. However, a good performance is found with the MPF, it appears to recover quickly after failing to detect the attractor of the system. On the other hand, the SIR filter (lower panels of Fig. \ref{ass_mpf_zobs-20npa}) looses track and it is unable to recover the true trajectory for a large number of cycles.

\end{document}